\documentclass[10pt,twocolumn]{article}

\usepackage[T1]{fontenc}
\usepackage[utf8]{inputenc}
\usepackage{geometry}
\usepackage{amsmath,amssymb,amsfonts}
\usepackage{bm}
\usepackage{graphicx}
\usepackage{booktabs}
\usepackage{tabularx}
\usepackage{multirow}
\usepackage{array}
\usepackage{float}
\usepackage{xcolor}
\usepackage{url}
\usepackage{makecell}

\newcommand{\authorbio}[3]{%
    \par\vspace{0.8em}
    \noindent
    \begin{minipage}[t]{0.20\linewidth}
        \vspace{0pt}
        \centering
        \includegraphics[
            width=\linewidth,
            height=1.05in,
            keepaspectratio
        ]{#1}
    \end{minipage}\hfill
    \begin{minipage}[t]{0.77\linewidth}
        \vspace{0pt}
        \footnotesize
        \textbf{#2} #3
    \end{minipage}
    \par\vspace{0.8em}
}
\usepackage{caption}
\makeatletter
\renewcommand\section{\@startsection{section}{1}{\z@}%
  {1.2ex plus .4ex}{.6ex plus .2ex}%
  {\normalfont\bfseries\fontsize{11}{13}\selectfont}}
\renewcommand\subsection{\@startsection{subsection}{2}{\z@}%
  {1.0ex plus .3ex}{.4ex plus .2ex}%
  {\normalfont\bfseries\fontsize{9.5}{11.5}\selectfont}}
\renewcommand\subsubsection{\@startsection{subsubsection}{3}{\z@}%
  {1.0ex plus .3ex}{.4ex plus .2ex}%
  {\normalfont\itshape\fontsize{9.5}{11.5}\selectfont}}
\makeatother

\usepackage[numbers,sort&compress,square]{natbib}

\usepackage{fancyhdr}
\usepackage[colorlinks=true,linkcolor=blue!70!black,%
             citecolor=blue!70!black,urlcolor=blue!70!black]{hyperref}

\newcommand{\Figref}[1]{Fig.~\ref{#1}}
\newcommand{\Tabref}[1]{Table~\ref{#1}}

\begin{document}
\raggedbottom

%-------------------------------------------------------------------------
%  FRONT MATTER
%-------------------------------------------------------------------------
\makeatletter
\twocolumn[%
\begin{@twocolumnfalse}
\vspace*{-1ex}

\noindent\textbf{Research Article}\par
\vspace{0.6ex}

{\large\bfseries SSP-DMGTimeNet: Physics-Constrained Learning for Spatiotemporal Trajectory Prediction of Vehicle Platoons\par}
\vspace{2ex}

\noindent
Yuhang Wang\textsuperscript{1,2},
Kailang Ma\textsuperscript{1},
Zirui Li\textsuperscript{3},
Mingfeng Fan\textsuperscript{4},
Kitae Jang\textsuperscript{1},
Changju Lee\textsuperscript{1,*},
Heye Huang\textsuperscript{1,*}\par

\vspace{1.5ex}

{\footnotesize
\textsuperscript{1}Cho Chun Shik Graduate School of Mobility, Korea Advanced Institute of Science \& Technology (KAIST), Daejeon 34051, Republic of Korea.\par\vspace{0.45ex}

\textsuperscript{2}Chinese Academy of Sciences, Beijing 100864, China.\par\vspace{0.45ex}

\textsuperscript{3}School of Mechanical and Aerospace Engineering, Nanyang Technological University, Singapore 639798, Singapore.\par\vspace{0.45ex}

\textsuperscript{4}Department of Mechanical Engineering, National University of Singapore, Singapore 119077, Singapore.\par\vspace{0.45ex}

\textsuperscript{*}Corresponding author. E-mail: [changju.lee@kaist.ac.kr;heye.huang@kaist.ac.kr]\par\vspace{0.45ex}

}

\vspace{1.8ex}

\noindent\textbf{ABSTRACT:} Existing car-following prediction methods mainly optimize trajectory accuracy, while rarely considering whether predicted disturbances propagate realistically along a vehicle platoon. This limitation may lead to accurate but string-unstable predictions. We propose SSP-DMGTimeNet, a physics-constrained learning framework for spatiotemporal trajectory prediction of vehicle platoons. The model combines multi-scale temporal representations with cross-vehicle interaction features to capture complex and time-varying platoon dynamics. A propagation-delay-aware causal attention mechanism explicitly models upstream-to-downstream disturbance propagation by learning response delays between adjacent vehicles and accumulating them along the platoon. In addition, time- and frequency-domain string-stability losses relieve disturbance amplification across both adjacent vehicles and arbitrary sub-platoons during training. Experiments on HighD show that SSP-DMGTimeNet achieves an unstable-window rate of 0.65\% for five-vehicle platoons and a maximum head-to-tail amplification of 0.898 on the ground-truth excitation subset, while maintaining competitive trajectory prediction performance. In zero-shot evaluation on NGSIM US-101 and I-80, the model achieves velocity MAEs of 1.316~m/s and 1.252~m/s, with unstable-window rates of 3.90\% and 4.10\%, respectively. These results demonstrate that incorporating platoon-level physical constraints can effectively balance trajectory prediction accuracy and disturbance propagation stability.\par
\vspace{1ex}

\par

\vspace{2ex}
\end{@twocolumnfalse}
]
\makeatother

%-------------------------------------------------------------------------
%  MAIN TEXT
%-------------------------------------------------------------------------

\section{Introduction}

Car-following prediction is fundamental to traffic flow modeling, driving safety assessment, and autonomous driving decision-making. Accurately predicting how a vehicle responds to the longitudinal motion of its preceding vehicle is important not only for realistic microscopic traffic simulation, but also for the design and validation of longitudinal control systems such as adaptive cruise control (ACC) and cooperative adaptive cruise control (CACC)~\cite{wang2023distributedmpc,bouadi2024multianticipation}

Early car-following studies primarily relied on analytical models such as the Intelligent Driver Model (IDM), Optimal Velocity Model (OVM), and Full Velocity Difference Model (FVDM). These models describe longitudinal driving dynamics using a small number of physically interpretable parameters, but their predefined functional forms often have limited capability to capture nonlinear and non-stationary behaviors in real traffic. Data-driven approaches have therefore been increasingly adopted. Recurrent neural networks, including Long Short-Term Memory (LSTM) and Gated Recurrent Unit (GRU), capture temporal dependencies in vehicle trajectories~\cite{huang2020probabilistic,liu2022probabilistic}, while interaction-aware extensions incorporate preceding-vehicle information to model inter-vehicle coupling. More recently, Transformer-based models have improved long-range temporal interaction modeling~\citep{giuliari2021transformer}, and graph-based methods have extended prediction from local vehicle pairs to larger multi-vehicle interaction structures. These approaches have achieved increasingly low trajectory prediction errors on datasets such as HighD and NGSIM.

Despite these advances, most existing models are optimized primarily for pointwise prediction accuracy at the individual-vehicle and individual-time-step levels~\cite{huang2026cogdrive,chen2025unveiling}. Even recent hybrid approaches that combine analytical car-following models with neural networks mainly use physical information to improve local trajectory prediction. As a result, the physical process by which disturbances propagate through a sequence of vehicles is rarely modeled as an explicit learning objective. A prediction model can therefore achieve a low MAE or RMSE while still producing unrealistic amplification of disturbances along a vehicle platoon. This reveals an important gap between trajectory accuracy and platoon-level physical consistency.

\begin{figure*}[t]
    \centering
    \includegraphics[width=\textwidth]{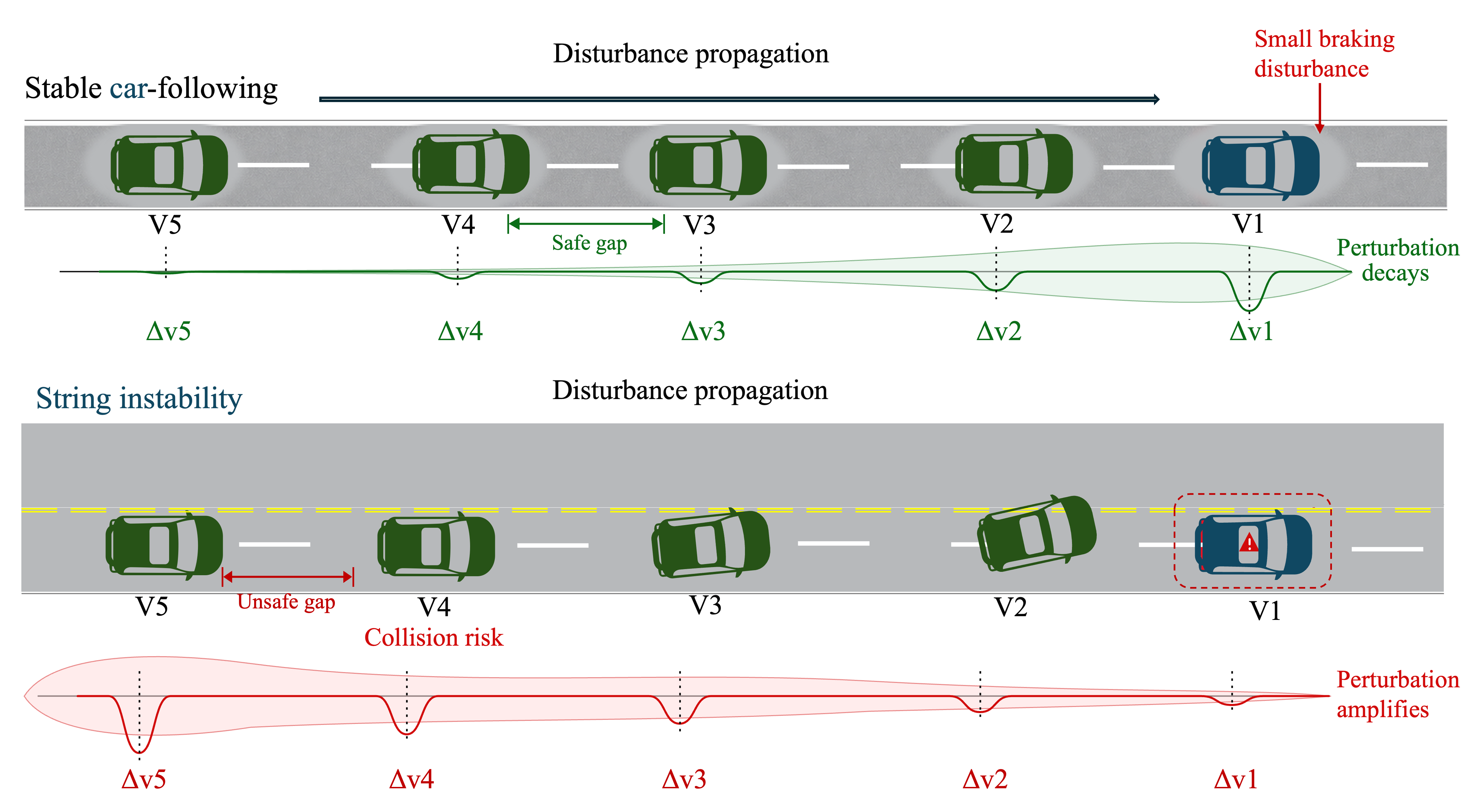}
    \caption{Illustration of disturbance propagation in a vehicle platoon.
    A small braking disturbance is introduced by the leading vehicle
    $\mathrm{V}_1$ and propagates backward to the following vehicles.
    In a string-stable platoon, the disturbance-response amplitude attenuates
    along the vehicle chain; in a string-unstable platoon, it is progressively
    amplified, potentially resulting in larger speed fluctuations and reduced
    inter-vehicle gaps.}
    \label{fig:string_stability}
\end{figure*}

String stability provides a natural physical criterion for addressing this gap. It characterizes whether disturbances are attenuated or amplified as they propagate downstream through successive vehicles. As illustrated in Fig.~\ref{fig:string_stability}A string-unstable platoon can amplify a small disturbance generated by an upstream vehicle, leading to increasingly pronounced speed oscillations and potentially degraded traffic safety. Empirical studies have shown that commercially implemented ACC systems may exhibit such string-unstable behavior, with upstream disturbances amplified after propagating through multiple vehicles~\citep{gunter2021commercialacc}. This issue is equally important for prediction: a model that reproduces or even exaggerates disturbance amplification may provide physically inconsistent future trajectories to downstream planning and control, even when its pointwise prediction error is small. However, existing trajectory prediction methods generally lack a unified framework that explicitly models disturbance propagation direction and response delay while directly constraining disturbance amplification during training.

To address this limitation, we propose SSP-DMGTimeNet, a physics-constrained learning framework for spatiotemporal trajectory prediction of vehicle platoons. Given 5\,s of historical states from consecutive vehicles traveling in the same lane, the model jointly predicts their longitudinal states over the next 3\,s. SSP-DMGTimeNet combines multi-scale temporal modeling with cross-vehicle equilibrium representations to capture both vehicle dynamics and inter-vehicle interactions. A sequence propagation delay-aware causal attention (SP-DACA) mechanism further models upstream-to-downstream disturbance propagation by learning response delays between adjacent vehicles and accumulating these delays along the vehicle chain. In addition, string-stability constraints are imposed in both the time and frequency domains to directly penalize disturbance amplification across neighboring vehicles and longer sub-platoons. In this way, physical disturbance propagation is incorporated into the learning process rather than evaluated only after prediction.
The main contributions of this work are summarized as follows:

\begin{itemize}

    \item We define a platoon-level physics-constrained spatiotemporal prediction problem that extends conventional vehicle- or pair-level car-following prediction to consecutive vehicle platoons, jointly considering trajectory accuracy and platoon-level disturbance propagation.

    \item We present SSP-DMGTimeNet, which integrates multi-scale temporal modeling, cross-vehicle cointegration features, and SP-DACA to explicitly capture delayed upstream-to-downstream disturbance propagation along vehicle platoons.

    \item We introduce time- and frequency-domain string-stability regularization over adjacent vehicles and arbitrary sub-platoons, converting disturbance amplification from a post-hoc evaluation metric into a training-time physical constraint. Extensive experiments on HighD and zero-shot evaluation on NGSIM validate the effectiveness and generalization of the proposed framework.

\end{itemize}

% Related-work section. Citation keys are synchronized with related_work_references.bib.

\section{Related Work}

\subsection{Vehicle Platoon Trajectory Prediction}

Classical car-following models describe longitudinal vehicle dynamics using analytical formulations with a small number of physically interpretable parameters. Representative models include the Intelligent Driver Model (IDM)~\cite{treiber2000congested}, Optimal Velocity Model (OVM)~\cite{bando1995dynamical}, and Full Velocity Difference Model (FVDM)~\cite{gong2008asymmetric}. These models provide good interpretability and calibratability, while delayed extensions further reveal the influence of driver response time on traffic-flow stability~\cite{bando1998explicitdelay}. However, their predefined functional forms can be restrictive when representing nonlinear and non-stationary driving behaviors.

Data-driven approaches provide greater flexibility for trajectory prediction. Recurrent architectures such as LSTM and GRU capture temporal dependencies in vehicle motion~\cite{hochreiter1997lstm,cho2014rnn}, while interaction-aware models incorporate preceding or neighboring vehicles to represent inter-vehicle coupling~\cite{altche2017lstm,chen2026respond,deo2018maneuverlstm}. Attention- and graph-based methods further extend interaction modeling to longer temporal horizons and larger multi-agent contexts. Representative approaches include Wayformer~\cite{nayakanti2023wayformer}, QCNet~\cite{zhou2023qcnet}, MTR++~\cite{shi2024mtrpp}, FJMP~\cite{rowe2023fjmp}, EqMotion~\cite{xu2023eqmotion}, and TSGN~\cite{wu2023tsgn}. More recently, generative and pre-trained models have explored diffusion-based prediction, sequential motion generation, masked pre-training, and large-scale trajectory representation learning~\cite{jiang2023motiondiffuser,zhu2025unitraj}.

These advances have substantially improved trajectory prediction on widely used datasets~\cite{huang2026knowledge}, such as HighD~\cite{krajewski2018highd} and NGSIM~\cite{fhwa2016ngsim}, with benchmarks such as FollowNet enabling systematic comparison of car-following models~\cite{chen2023follownet}. However, most methods remain optimized primarily for vehicle-level pointwise errors. Prediction accuracy does not necessarily imply physical consistency or traffic-flow stability~\cite{chen2024datadriven}, and even hybrid physics--data approaches generally focus on improving individual-vehicle predictions~\cite{bhattacharyya2022hybrid}. Consequently, disturbance propagation across a complete vehicle platoon has rarely been treated as an explicit learning objective.

\subsection{String Stability and Cooperative Adaptive Cruise Control}

String stability characterizes whether disturbances are attenuated or amplified as they propagate downstream through a vehicle platoon. Classical studies established stability conditions for interconnected vehicle strings and extended them from linear systems to more general cascaded systems~\cite{swaroop1996string,ploeg2014lp}. Subsequent work investigated traffic-flow stability in connected and automated traffic~\cite{talebpour2016influence,qin2023mixedstability}, while field experiments demonstrated that appropriately controlled automated vehicles can suppress stop-and-go waves~\cite{stern2018dissipation}.

Importantly, commercially deployed ACC systems are not necessarily string stable. Experimental studies have characterized their dynamic responses and traffic-flow impacts~\cite{makridis2020empiricalacc,makridis2021openacc}, and real-vehicle tests have shown that some production ACC systems amplify upstream speed disturbances~\cite{gunter2021commercialacc}. CACC seeks to mitigate this problem through inter-vehicle information sharing, and existing studies have examined its stability under communication effects, mixed traffic, distributed control, and time delays~\cite{wang2023distributedmpc,bouadi2024multianticipation}.

Most of this literature treats string stability as a property of vehicle dynamics or control systems. In trajectory prediction, however, it is typically evaluated only after trajectories have been generated, if considered at all. This motivates incorporating disturbance amplification directly into the learning objective so that predicted trajectories preserve platoon-level physical consistency rather than only pointwise accuracy.

\subsection{Propagation Delay and Causal Attention}

Propagation delay is a key mechanism underlying inter-vehicle disturbance transmission. Driver response, vehicle actuation, and communication delays cause downstream vehicles to react to earlier states of upstream vehicles rather than to synchronized observations. Delayed car-following models have shown that such response delays can alter stability boundaries and induce traffic oscillations~\cite{bando1998explicitdelay,orosz2010trafficjams}. Explicitly representing delay is therefore important for modeling disturbance propagation within vehicle platoons.

Modern temporal models capture causal dependencies using causal convolutions, masked attention, and multi-scale temporal representations~\cite{bai2018empirical,liu2019integrated}. Multi-agent predictors further combine attention with social and scene interactions~\cite{yuan2021agentformer,ngiam2022scene}. However, temporal causality in these models mainly prevents access to future observations; it does not explicitly represent how response delays accumulate with topological distance along a vehicle platoon. Likewise, attention weights do not guarantee physically meaningful disturbance propagation or string-stable predictions.

Recent diffusion and masked trajectory models improve multimodal motion forecasting~\cite{jiang2023motiondiffuser,capellera2025uncertaintydiffusion}, but their objectives remain centered on future trajectory distributions rather than platoon-level stability. In contrast, SSP-DMGTimeNet jointly models learnable propagation delays, directional upstream-to-downstream interactions, and string-stability constraints, allowing disturbance propagation to be explicitly represented and constrained during trajectory prediction.

\section{Problem Definition}

\subsection{Platoon-Level Prediction Task}

Our objective is to characterize the future state of a vehicle platoon at the prediction level. The model should not only predict the future trajectory of each vehicle, but also determine whether disturbances are amplified as they propagate downstream. We therefore first define the input and output of the platoon-level prediction task.

Consider a sequence of consecutive vehicles traveling in the same lane,
$C_1 \to C_2 \to \cdots \to C_N$, where $C_1$ denotes the leading vehicle and $C_i$ ($i=2,\ldots,N$) denotes the $i$th following vehicle. Given a historical observation window of $T$ frames, the model predicts the longitudinal states of the entire platoon over a future horizon of $P$ frames. The input and output are represented as

\begin{equation}
X \in \mathbb{R}^{B \times T \times N \times F},
\qquad
\hat{Y} \in \mathbb{R}^{B \times P \times N \times D},
\label{eq:io_tensor}
\end{equation}
where $B$ is the batch size, $F$ is the number of input features per vehicle, and $D$ is the number of predicted variables per vehicle. The input contains $F=8$ features: absolute longitudinal position, position relative to the leading vehicle, velocity, acceleration, inter-vehicle spacing, relative velocity, relative acceleration, and time headway. Relative quantities are included to reduce sensitivity to absolute road-position differences.

The output contains $D=4$ variables: velocity $\hat{v}_i$, inter-vehicle spacing $\hat{s}_i$, acceleration $\hat{a}_i$, and position relative to the leading vehicle $\widehat{(x_i-x_1)}$. These variables support four complementary aspects of platoon-level evaluation: string stability, safety and spacing consistency, comfort and energy-related dynamics, and kinematic consistency.

\subsection{Prediction-Level String-Stability Criterion}

Pointwise trajectory errors, such as MAE, do not indicate whether disturbances are amplified as they propagate through a vehicle platoon. To capture this platoon-level physical property, we adapt the classical concept of string stability into finite-horizon criteria that can be evaluated directly on predicted trajectories.

For the predicted velocity sequence $\hat{v}_i$, we first define the detrended component as

\begin{equation}
\tilde{v}_i = \hat{v}_i -\mathrm{MA}_{W} (\hat{v}_i),
\label{eq:detrend}
\end{equation}
where $\mathrm{MA}_{W}(\cdot)$ denotes a moving-average operator with window length $W$. This operation removes low-frequency trends while retaining disturbance-related fluctuations.

For an arbitrary sub-platoon from $C_j$ to $C_i$, $j<i$, the time-domain head-to-tail amplification factor is defined as

\begin{equation}
A_{j \to i}
=
\frac{\left\|\tilde{v}_i\right\|_2}
{\left\|\tilde{v}_j\right\|_2+\epsilon},
\qquad j<i,
\label{eq:amp_factor}
\end{equation}
where $\epsilon$ is a small constant for numerical stability. The adjacent-vehicle amplification factor,
$A_i \triangleq A_{i-1 \to i}$, is a special case of this definition.

The time-domain ratio characterizes the overall amplification of a disturbance but does not reveal frequency-selective amplification. We therefore apply a zero-padded discrete Fourier transform to $\tilde{v}_i$ and denote the resulting spectrum by $V_i(f)$. The frequency-domain transfer gain over the traffic-disturbance frequency band $\mathcal{F}$ is defined as

\begin{equation}
G_{j \to i}(f)
=
\frac{\left|V_i(f)\right|}
{\left|V_j(f)\right|+\epsilon},
\qquad
f \in \mathcal{F}.
\label{eq:freq_gain}
\end{equation}

The two measures characterize complementary aspects of disturbance propagation. $A_{j \to i}$ captures the overall amplification across a sub-platoon, whereas $G_{j \to i}(f)$ identifies frequency-dependent amplification that may arise in oscillatory traffic conditions.

Given a tolerance $\delta \geq 0$, a prediction window is defined as $\delta$-string stable if

\begin{equation}
\max_{j<i} A_{j \to i}
\leq 1+\delta,
\qquad
\max_{f \in \mathcal{F}}
\max_{j<i}
G_{j \to i}(f)
\leq 1+\delta.
\label{eq:stable_cond}
\end{equation}

Otherwise, the window is classified as unstable. To quantify the accumulated severity of time-domain amplification beyond the stability boundary, we further define the exceedance area as

\begin{equation}
E
=
\sum_{j<i}
\max
\left(
0,\,
A_{j \to i}-1-\delta
\right).
\label{eq:exceed_area}
\end{equation}

A smaller $E$ therefore indicates weaker cumulative disturbance amplification across the vehicle platoon.

\begin{figure*}
    \centering
    \includegraphics[width=0.9\linewidth]{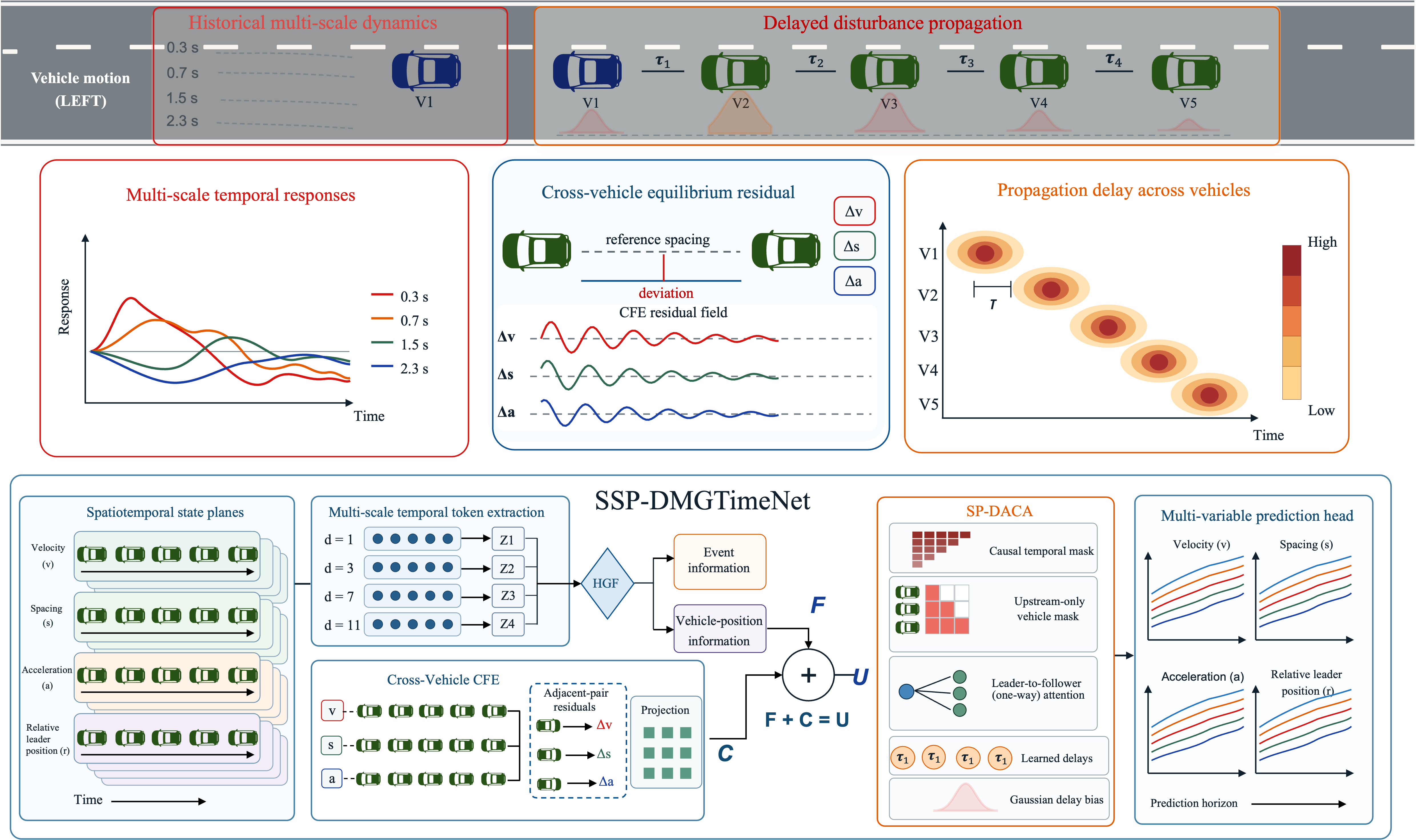}
    \caption{Overview of SSP-DMGTimeNet. The framework consists of four main components: multi-scale temporal token extraction, Hierarchical Gated Fusion (HGF), Cross-Vehicle Cointegration Feature Extraction (CFE), and Sequence Propagation Delay-Aware Causal Attention (SP-DACA).}
    \label{fig:framework}
\end{figure*}

\section{Method}

\subsection{Framework}

As shown in \Figref{fig:framework}, SSP-DMGTimeNet takes the historical states of a five-vehicle platoon as input and jointly predicts the future longitudinal states of all vehicles. Historical dynamics are first encoded at four temporal scales with effective receptive fields of approximately 0.3--2.3\,s. The model contains two parallel feature-extraction branches. The main branch projects normalized vehicle states into a shared hidden space and extracts multi-scale temporal tokens using temporal convolutions with different receptive fields. Hierarchical Gated Fusion (HGF) then adaptively combines these representations according to the current motion state and vehicle position within the platoon. In parallel, Cross-Vehicle Cointegration Feature Extraction (CFE) derives residual features from the velocity, spacing, and acceleration of adjacent vehicles to characterize deviations from inter-vehicle equilibrium. The two branches are fused by element-wise addition.

The fused representation is then processed by Sequence Propagation Delay-Aware Causal Attention (SP-DACA), which models the platoon as a directed upstream-to-downstream chain using learnable response delays, Gaussian delay biases, and a joint temporal--vehicle causal mask. After stacked SP-DACA layers, a multi-variable prediction head outputs the velocity, spacing, acceleration, and position relative to the leading vehicle over the prediction horizon.

\subsection{Multi-Scale Temporal Token Extraction}

Vehicle dynamics exhibit distinct characteristics across temporal scales. Short time scales capture transient changes such as rapid acceleration or braking, whereas longer scales characterize persistent motion trends. To represent both types of dynamics, the historical observations are first projected into a $D$-dimensional hidden space,
$X \in \mathbb{R}^{T \times N \times D}$.
We define $M=4$ temporal scales,
$\mathcal{S}=\{0.4,\,0.8,\,1.6,\,2.4\}$\,s.
For the $m$th scale, the dilation rate is determined by

\begin{equation}
d_m =
\max\left(
1,\,
\left\lfloor
\frac{\mathrm{round}(s_m f_s)-1}{2}
\right\rfloor
\right),
\label{eq:dilation}
\end{equation}
where $s_m$ denotes the $m$th temporal scale and $f_s$ is the sampling frequency. At $f_s=10$\,Hz, the resulting dilation rates are
$\{1,\,3,\,7,\,11\}$.

For each vehicle, the temporal sequence is independently processed using a depthwise dilated convolution with a kernel size of 3, followed by a pointwise convolution for channel-wise feature interaction:

\begin{equation}
\begin{aligned}
Z^{(m)}
={}&
\mathrm{Dropout}\Bigl(
\mathrm{GELU}\bigl(
\mathrm{LN}\bigl[
\\[-3pt]
&\quad
\mathrm{PWConv}^{(m)}
(
\mathrm{DWConv}^{(m)}_{d_m}(X)
)
\bigr]
\bigr)
\Bigr).
\end{aligned}
\label{eq:multiscale}
\end{equation}
where $\mathrm{DWConv}^{(m)}_{d_m}$ denotes the depthwise dilated convolution at the $m$th temporal scale with dilation rate $d_m$, and $\mathrm{PWConv}^{(m)}$ denotes the pointwise convolution for feature mixing across channels. The resulting representation $Z^{(m)} \in \mathbb{R}^{T \times N \times D}$ defines the temporal tokens at scale $m$.

Symmetric padding preserves the original temporal length. The four branches have effective receptive fields of 3, 7, 15, and 23 frames, corresponding to approximately 0.3, 0.7, 1.5, and 2.3\,s, respectively. The module therefore produces

\begin{equation}
\mathcal{Z}
=
\left\{
Z^{(1)},
Z^{(2)},
Z^{(3)},
Z^{(4)}
\right\}.
\label{eq:multi_token}
\end{equation}

At this stage, temporal dynamics are modeled independently for each vehicle, without cross-vehicle interaction. The resulting multi-scale temporal tokens are subsequently passed to HGF for adaptive fusion.

\subsection{HGF: Hierarchical Gated Fusion}

Different temporal scales are informative under different driving conditions. During steady cruising, longer-scale representations better capture gradual motion trends, whereas rapid braking or acceleration is more effectively represented at shorter scales. Moreover, the temporal characteristics of a disturbance may change as it propagates downstream, making a fixed combination of temporal scales suboptimal across vehicles and time.

We therefore introduce Hierarchical Gated Fusion (HGF), which assigns adaptive scale weights to each vehicle at each time step. The gating input combines the finest-scale representation with the vehicle's position within the platoon:

\begin{equation}
h_{t,i}
=
\left[
Z^{(1)}_{t,i,:}
\,\Vert\,
e_i
\right]
\in
\mathbb{R}^{D+N},
\label{eq:hgf_input}
\end{equation}
where $e_i \in \{0,1\}^{N}$ is the one-hot encoding of the position of vehicle $i$. The finest-scale representation provides sensitivity to transient motion changes, while the positional encoding allows different vehicles along the platoon to adopt different temporal-scale preferences.

The gating vector is mapped through a two-layer multilayer perceptron with GELU activation and normalized using softmax:

\begin{equation}
g_{t,i}
=
\mathrm{softmax}
\left(
W_2
\sigma
\left(
W_1 h_{t,i}+b_1
\right)
+b_2
\right)
\in
\Delta^{M-1},
\label{eq:hgf_gate}
\end{equation}
where $\sigma(\cdot)$ denotes GELU, $W_1$, $W_2$, $b_1$, and $b_2$ are learnable parameters, and $\Delta^{M-1}$ denotes the $(M-1)$-dimensional probability simplex. The fused representation is then obtained as

\begin{equation}
F_{t,i,:}
=
\mathrm{LN}
\left(
\sum_{m=1}^{M}
g_{t,i}^{(m)}
Z_{t,i,:}^{(m)}
\right),
\qquad
F \in
\mathbb{R}^{T \times N \times D}.
\label{eq:hgf_fusion}
\end{equation}

The softmax normalization ensures
$g_{t,i}^{(m)} \geq 0$
and
$\sum_{m=1}^{M} g_{t,i}^{(m)}=1$,
so that the fused feature forms a convex combination of the temporal-scale representations.
The same $M$ gating coefficients are shared across all $D$ feature channels for each vehicle--time pair. This design keeps the gating mechanism compact and directly interpretable: $g_{t,i}$ indicates the relative preference of vehicle $i$ at time $t$ for each temporal scale. The fused representation $F$ is subsequently combined with the cross-vehicle cointegration features before being passed to SP-DACA.

\subsection{Cross-Vehicle Cointegration Feature Extraction}

Individual vehicle trajectories are often non-stationary, with large-scale trends dominating the relatively small fluctuations associated with car-following interactions. In contrast, adjacent vehicles tend to exhibit a more stable equilibrium relationship. Motivated by cointegration, where a linear combination of non-stationary series can remain stationary, we use the resulting residual to characterize instantaneous deviations from the long-term inter-vehicle equilibrium. For each channel, the cross-vehicle equilibrium residual is defined as
\begin{equation}
r_{t,i}^{(c)}
=
o_{t,i}^{(c)}
-
\alpha_{i-1}^{(c)} o_{t,i-1}^{(c)}
-
\beta_{i-1}^{(c)},
\qquad
r_{t,1}^{(c)} \equiv 0,
\qquad
\mu_i=\mathbb{1}[i\geq 2],
\label{eq:cfe_residual}
\end{equation}
where $o_{t,i}^{(c)}$ denotes the raw observation of vehicle $i$ at time $t$, and $c\in\{v,s,a\}$ corresponds to velocity, spacing, and acceleration, respectively. The parameters $\alpha_{i-1}^{(c)}$ and $\beta_{i-1}^{(c)}$ represent the equilibrium coefficient and offset for the adjacent pair $(i-1,i)$, while $\mu_i$ masks the leading vehicle, which has no upstream reference. This residual suppresses common platoon-level trends and becomes prominent when a vehicle deviates from its local car-following equilibrium.

To maintain a physically meaningful range and avoid degenerate solutions, the equilibrium coefficient is bounded through reparameterization:
\begin{equation}
\alpha_j^{(c)}
=
\alpha_{\min}
+
\left(
\alpha_{\max}-\alpha_{\min}
\right)
\mathrm{sigmoid}
\left(
\hat{\alpha}_j^{(c)}
\right),
\label{eq:alpha_reparam}
\end{equation}
where $\hat{\alpha}_j^{(c)}$ is an unconstrained learnable parameter and $j=1,\ldots,N-1$ indexes adjacent vehicle pairs. We set $[\alpha_{\min},\alpha_{\max}]=[0.5,1.5]$. This bounded parameterization also prevents the residual from degenerating toward the raw observation as $\alpha\rightarrow 0$. The parameters are initialized with $\alpha=1$ and $\beta=0$, under which Eq.~(\ref{eq:cfe_residual}) reduces to a first-order spatial difference between adjacent vehicles.

The three-channel residual vector is projected into the hidden space and injected into the main branch through a residual connection:
\begin{equation}
C_{t,i}
=
\mu_i\,
\mathrm{LN}
\left(
W_2
\sigma
\left(
W_1 r_{t,i}+b_1
\right)
+b_2
\right),
\qquad
U=F+C,
\label{eq:cfe_output}
\end{equation}
where $r_{t,i}\in\mathbb{R}^{3}$ contains the residuals of velocity, spacing, and acceleration, $\sigma(\cdot)$ denotes GELU, $F$ is the HGF output, and $U$ is the fused representation passed to SP-DACA. The additive fusion preserves the backbone dimensionality and reduces to an identity mapping when the cross-vehicle residual approaches zero.

\subsection{SP-DACA: Sequence Propagation Delay-Aware Causal Attention}

\begin{figure*}[t]
    \centering
    \includegraphics[width=0.85\textwidth]{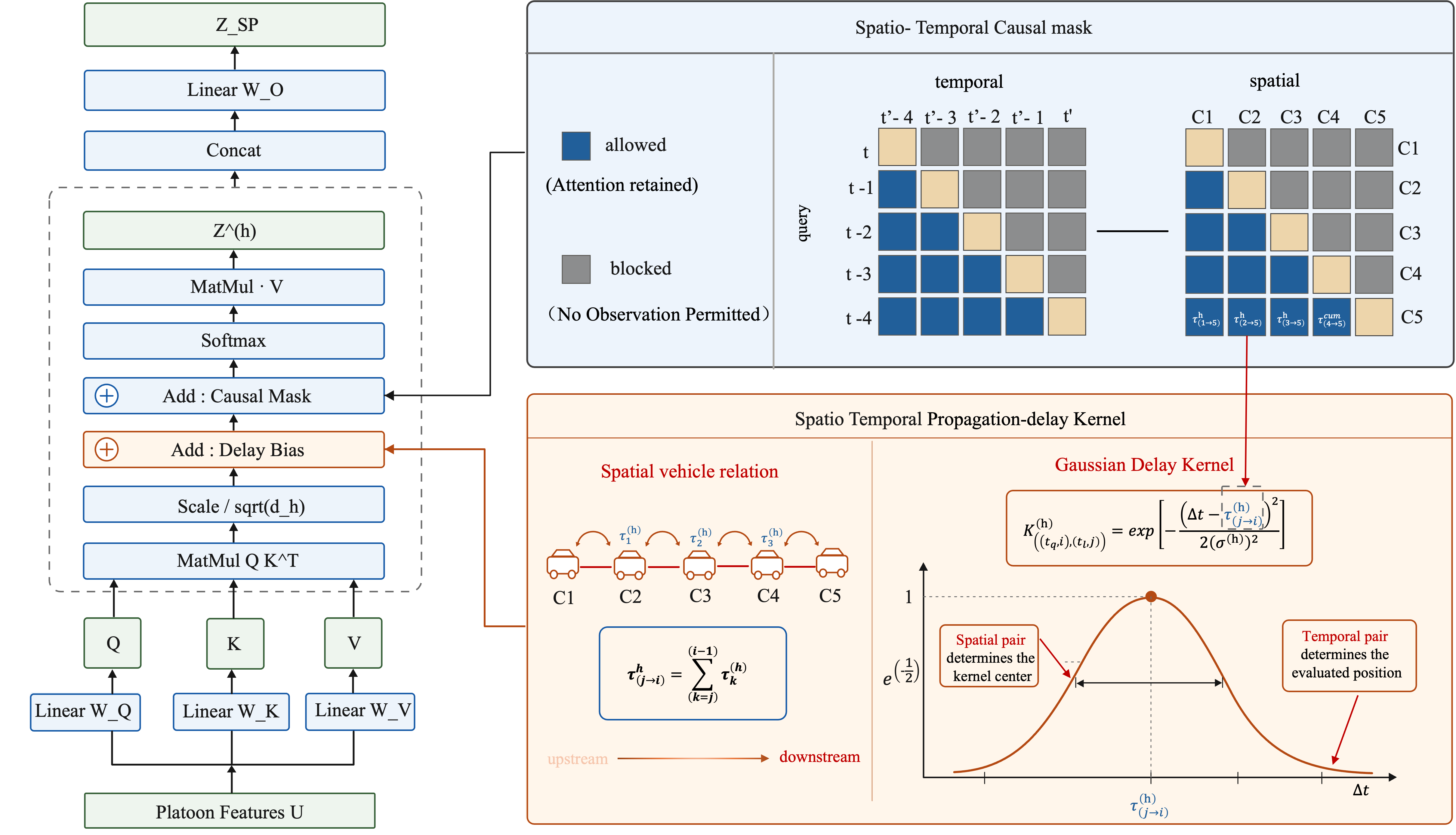}
    \caption{
    Architecture of the proposed SP-DACA, which integrates a learned propagation-delay bias with a joint temporal–vehicle causal mask. The delay-centered Gaussian kernel emphasizes upstream states consistent with the expected disturbance-propagation delay.
    }
    \label{fig:spdaca}
\end{figure*}

Disturbances in a vehicle platoon generally propagate from upstream to downstream through successive car-following responses. The motion of an upstream vehicle can influence its followers, whereas downstream states should not be used to explain the upstream response. Moreover, perception, decision-making, and vehicle actuation introduce response delays, such that a following vehicle reacts to an earlier state of its predecessor. Standard attention does not explicitly encode either this directional structure or the associated propagation delay. We therefore introduce Sequence Propagation Delay-Aware Causal Attention (SP-DACA).As shown in \Figref{fig:spdaca}

The input to SP-DACA is the fused representation from HGF and CFE:
\begin{equation}
U = F + C,
\qquad
U \in \mathbb{R}^{T\times N\times D}.
\label{eq:spdaca_input}
\end{equation}
The temporal and vehicle dimensions are flattened into $TN$ spatiotemporal tokens. For attention head $h$, the query, key, and value representations are obtained by
\begin{equation}
Q^{(h)}=UW_Q^{(h)},
\qquad
K^{(h)}=UW_K^{(h)},
\qquad
V^{(h)}=UW_V^{(h)},
\label{eq:qkv}
\end{equation}
and the standard content-based attention score is
\begin{equation}
S_{\mathrm{content}}^{(h)}
=
\frac{
Q^{(h)}K^{(h)\top}
}{
\sqrt{d_h}
},
\label{eq:content_score}
\end{equation}
where $d_h$ denotes the feature dimension of each attention head.

To represent the response delay between adjacent vehicles, each adjacent vehicle pair learns a propagation delay:
\begin{equation}
\tau_k^{(h)}
=
\tau_{\min}
+
\left(
\tau_{\max}-\tau_{\min}
\right)
\mathrm{sigmoid}
\left(
\hat{\tau}_k^{(h)}
\right),
\label{eq:tau_learn}
\end{equation}
where $k=1,\ldots,N-1$ indexes adjacent vehicle pairs and $\hat{\tau}_k^{(h)}$ is an unconstrained learnable parameter. The propagation delay is bounded within
$[\tau_{\min},\tau_{\max}]=[0.3,2.5]$\,s. For non-adjacent vehicles, the propagation delay is accumulated along the intermediate vehicle pairs:
\begin{equation}
\tau_{j\rightarrow i}^{(h)}
=
\sum_{k=j}^{i-1}
\tau_k^{(h)},
\qquad
j<i.
\label{eq:tau_cumsum}
\end{equation}
This additive formulation reflects the sequential nature of disturbance transmission through the vehicle platoon.

Given the accumulated delay, we construct a continuous Gaussian delay kernel centered at $\tau_{j\rightarrow i}^{(h)}$:
\begin{equation}
\mathcal{K}_{j\rightarrow i}^{(h)}(\Delta t)
=
\exp
\left[
-\frac{
\left(
\Delta t-\tau_{j\rightarrow i}^{(h)}
\right)^2
}{
2\left(\sigma^{(h)}\right)^2
}
\right],
\qquad
\Delta t
=
\frac{t_q-t_\ell}{f_s},
\label{eq:gauss_kernel}
\end{equation}
where $t_q$ and $t_\ell$ denote the query and key time indices, respectively, $f_s$ is the sampling frequency, and $\sigma^{(h)}$ controls the kernel width for attention head $h$. The kernel assigns larger values when the temporal separation $\Delta t$ is consistent with the expected propagation delay. The Gaussian kernel is subsequently converted into a logarithmic delay bias
$B_{(t_q,i),(t_\ell,j)}^{(h)}$.

The delay bias is added to the content-based attention score together with a joint temporal--vehicle causal mask:
\begin{equation}
A^{(h)}
=
\mathrm{softmax}
\left(
S_{\mathrm{content}}^{(h)}
+
\gamma^{(h)} B^{(h)}
+
M
\right),
\label{eq:attn_score}
\end{equation}
where $\gamma^{(h)}$ is a learnable scaling coefficient for the delay bias. The joint causal mask is defined as
\begin{equation}
M_{(t_q,i),(t_\ell,j)}
=
\begin{cases}
0,
& t_\ell \leq t_q \ \text{and}\ j\leq i,\\
-\infty,
& \text{otherwise}.
\end{cases}
\label{eq:mask}
\end{equation}
The condition $t_\ell\leq t_q$ prevents access to future observations, while $j\leq i$ restricts vehicle $i$ to attending only to itself and upstream vehicles. The resulting attention output is
\begin{equation}
O^{(h)}
=
A^{(h)}V^{(h)}.
\label{eq:attn_output}
\end{equation}
Outputs from all attention heads are concatenated and linearly projected back to
$\mathbb{R}^{T\times N\times D}$, followed by residual connections and a feed-forward network. We stack three SP-DACA layers to progressively capture delayed disturbance propagation across the vehicle platoon.

\subsection{String-Stability Loss and Overall Objective}

To suppress disturbance amplification along the vehicle platoon, we introduce string-stability losses at three complementary levels: adjacent vehicles, arbitrary sub-platoons, and the frequency domain. We first define the exceedance penalty as
\begin{equation}
\phi_\delta(x)
=
\left[
\max
\left(
0,\,
x-1-\delta
\right)
\right]^2,
\label{eq:phi}
\end{equation}
where $\delta\geq0$ denotes the stability tolerance. Based on the time-domain amplification factor and frequency-domain transfer gain defined in Section~3, the three stability losses are
\begin{equation}
L_{\mathrm{adj}}
=
\sum_{i=2}^{N}
\phi_\delta
\left(
A_{i-1\rightarrow i}
\right),
\label{eq:loss_adj}
\end{equation}
\begin{equation}
L_{\mathrm{sub}}
=
\sum_{1\leq j<i\leq N}
\phi_\delta
\left(
A_{j\rightarrow i}
\right),
\label{eq:loss_sub}
\end{equation}
\begin{equation}
L_{\mathrm{fft}}
=
\sum_{1\leq j<i\leq N}
\sum_{f\in\mathcal{F}}
\phi_\delta
\left(
G_{j\rightarrow i}(f)
\right).
\label{eq:loss_fft}
\end{equation}
Here, $L_{\mathrm{adj}}$ penalizes local disturbance amplification between adjacent vehicles, while $L_{\mathrm{sub}}$ constrains head-to-tail amplification over arbitrary sub-platoons and prevents small local amplifications from accumulating along the vehicle chain. The frequency-domain term $L_{\mathrm{fft}}$ further suppresses frequency-selective amplification within the traffic-disturbance band $\mathcal{F}$.

The overall training objective combines the trajectory prediction loss with the three string-stability terms:
\begin{equation}
L
=
L_{\mathrm{pred}}
+
\lambda_1 L_{\mathrm{adj}}
+
\lambda_2 L_{\mathrm{sub}}
+
\lambda_3 L_{\mathrm{fft}},
\label{eq:loss_total}
\end{equation}
where $L_{\mathrm{pred}}$ denotes the prediction loss, and $\lambda_1$, $\lambda_2$, and $\lambda_3$ control the contributions of the adjacent-vehicle, sub-platoon, and frequency-domain stability constraints, respectively.

\section{Experiments}

\subsection{Datasets}

\textbf{HighD}. We use HighD as the primary dataset for training and evaluation. Recordings 01--45, 46--50, and 51--60 are used for training, validation, and testing, respectively. Vehicles are grouped by lane and ordered longitudinally to construct consecutive same-lane platoons $C_1,\ldots,C_N$. Samples are retained only when all vehicles are continuously observed, adjacent spacings satisfy $s_i>0$, and vehicle order remains unchanged throughout the observation and prediction windows. To emphasize non-stationary car-following dynamics, we retain segments whose leader velocity standard deviation and velocity drop are both above the within-recording median ($q=0.5$). Trajectories are low-pass filtered and downsampled to 10\,Hz. The main setting uses a 5\,s history, a 3\,s prediction horizon, a 1\,s sliding step, and five-vehicle platoons ($N=5$), yielding 62,183 training, 4,672 validation, and 1,847 test windows.

\textbf{NGSIM}. We use the US-101 and I-80 subsets of NGSIM for cross-dataset evaluation. Their trajectories are mapped to the same coordinate convention as HighD. For zero-shot testing, the HighD-trained model is directly evaluated on NGSIM without target-domain fine-tuning or re-estimation of normalization statistics. The Montanino--Punzo reconstructed trajectories from the I-80 0400--0415 period are additionally used to assess sensitivity to trajectory smoothing.

\subsection{Baselines and Evaluation Metrics}

We compare SSP-DMGTimeNet with ten baselines spanning three categories. Physics-based models include IDM, OVM, and FVDM. Data-driven models include LSTM, interaction-aware LSTM (Int-LSTM), Transformer, and Full-graph Attention. Hybrid physics--learning approaches include DMGTimeNet cascade and CNN-Int-LSTM-IDM. All learning-based models are trained and evaluated using the same five-vehicle platoon samples. AdamW is used for optimization with 80 epochs and a batch size of 64.

Prediction accuracy is evaluated using MAE and RMSE for velocity, spacing, and acceleration. Platoon-level stability is evaluated using the disturbance-amplification measures defined in Section~3, including the unstable-window rate, maximum amplification, and exceedance area. We additionally report tail-vehicle velocity error and RMS jerk to characterize downstream prediction accuracy and motion smoothness.

\subsection{Results on HighD}

\Tabref{tab:accuracy} summarizes the prediction results on the HighD test set. SSP-DMGTimeNet does not achieve the lowest velocity error, but remains competitive in spacing and acceleration prediction despite the additional physical and stability constraints. Its tail-vehicle velocity MAE is 0.222\,m/s, compared with an average velocity MAE of 0.347\,m/s across the platoon, indicating that prediction error does not progressively deteriorate toward the downstream end. This observation, however, does not by itself establish stable disturbance propagation, which is evaluated separately below.

\begin{table*}[t]
\centering
\caption{Trajectory prediction performance on the HighD test set.}
\label{tab:accuracy}
\small
\resizebox{\textwidth}{!}{%
\begin{tabular}{lccrrrrrr}
\toprule
Model &
$v$-MAE$\downarrow$ &
$v$-RMSE &
$s$-MAE &
$s$-RMSE &
$a$-MAE &
$a$-RMSE &
Tail $v$-MAE &
Unstable windows (\%)$^{*}$ \\
\midrule
Int-LSTM                     & \textbf{0.102} & \textbf{0.158} & \textbf{0.262} & \textbf{0.374} & \textbf{0.068} & \textbf{0.115} & \textbf{0.114} & 91.30 \\
Transformer                  & 0.158 & 0.215 & 0.399 & 0.544 & 0.079 & 0.128 & 0.170 & 85.71 \\
Full-graph Attention         & 0.164 & 0.225 & 0.306 & 0.447 & 0.081 & 0.130 & 0.169 & 84.62 \\
LSTM                         & 0.213 & 0.282 & 0.511 & 0.728 & 0.074 & 0.118 & 0.209 & 95.12 \\
IDM cascade$^{\dagger}$      & 0.366 & 0.543 & 0.446 & 0.733 & 0.315 & 0.446 & 0.365 & --- \\
CNN-Int-LSTM-IDM             & 0.564 & 0.815 & 0.633 & 1.030 & 0.254 & 0.561 & 0.543 & 88.89 \\
DMGTimeNet cascade$^{\dagger}$
                             & 0.184 & 0.285 & 0.418 & 0.592 & 0.107 & 0.169 & 0.173 & --- \\
OVM cascade$^{\dagger}$      & 0.722 & 0.996 & 0.512 & 0.863 & 0.450 & 0.555 & 0.744 & --- \\
FVDM cascade$^{\dagger}$     & 0.758 & 1.024 & 0.519 & 0.867 & 0.478 & 0.599 & 0.817 & --- \\
\textbf{SSP-DMGTimeNet (ours)}
                             & 0.347 & 0.513 & 0.290 & 0.412 & 0.077 & 0.126 & 0.222 & \textbf{0.65} \\
\bottomrule
\end{tabular}%
}
\\[2pt]
{\footnotesize
$^{*}$ Computed only for windows with predicted-leader detrended RMS $\geq 0.05$\,m/s.
$^{\dagger}$ The physics-based/cascade baselines use a constant-speed leader, for which the stability metrics are undefined.}
\end{table*}

The stability results are reported in \Tabref{tab:stability}. Under the per-model excitation floor, SSP-DMGTimeNet achieves an unstable-window rate of only 0.65\% and retains valid leader excitation in all 1,847 test windows. In contrast, the other learning-based models retain only 18--41 windows because their predicted leader trajectories are frequently over-smoothed. This difference in valid support makes direct stability comparison under the per-model criterion potentially biased.

We therefore additionally evaluate all models against the 62 windows in which the ground-truth leader exhibits valid excitation. SSP-DMGTimeNet covers all 62 windows, with an unstable-window rate of 0\% and a maximum head-to-tail amplification of 0.898, below the string-stability boundary of one. In contrast, the other learning-based models retain only 11--24 valid windows and exhibit unstable-window rates of 83.33--100\%, with maximum amplification exceeding 2.7. The ground-truth trajectories themselves are unstable in 93.55\% of these windows, with a maximum amplification of 9.683. These results indicate that SSP-DMGTimeNet suppresses downstream disturbance amplification rather than simply reproducing the unstable behavior observed in human-driven platoons.

The improved stability is not obtained through excessively aggressive motion. SSP-DMGTimeNet achieves a 5th-percentile TTC of 12.175\,s and an RMS jerk of 0.116, both within the range of the learning-based baselines, indicating that disturbance attenuation is achieved while maintaining reasonable safety margins and motion smoothness.

\begin{table*}[t]
\centering
\caption{Platoon-level stability on the HighD test set under the per-model and GT-excitation criteria.}
\label{tab:stability}

\small
\renewcommand{\arraystretch}{1.08}

\begin{tabular*}{\textwidth}
{@{\extracolsep{\fill}}lcccccccc@{}}
\toprule
Model &
\makecell{Unstable\\(\%)$^{*}$} &
\makecell{Max.\\amp.$^{*}$} &
\makecell{Exceedance\\area$^{*}$} &
\makecell{Valid\\windows$^{*}$} &
\makecell{GT unstable\\(\%)} &
\makecell{GT max.\\amp.} &
\makecell{GT valid\\windows} &
\makecell{RMS\\jerk} \\
\midrule

Int-LSTM
& 91.30 & 2.865 & 12.794 & 23
& 88.89 & 2.865 & 18 & 0.125 \\

Transformer
& 85.71 & 2.735 & 20.277 & 28
& 83.33 & 2.735 & 24 & 0.107 \\

Full-graph Attention
& 84.62 & 3.529 & 15.752 & 26
& 86.36 & 3.529 & 22 & 0.106 \\

LSTM
& 95.12 & 3.240 & 11.625 & 41
& 100.00 & 3.240 & 11 & 0.121 \\

IDM cascade$^{\dagger}$
& --- & --- & --- & 0
& --- & --- & 0 & 0.572 \\

CNN-Int-LSTM-IDM
& 88.89 & 7.253 & 27.744 & 18
& 93.33 & 7.253 & 15 & 2.727 \\

DMGTimeNet cascade$^{\dagger}$
& --- & --- & --- & 0
& --- & --- & 0 & 0.101 \\

OVM cascade$^{\dagger}$
& --- & --- & --- & 0
& --- & --- & 0 & \textbf{0.030} \\

FVDM cascade$^{\dagger}$
& --- & --- & --- & 0
& --- & --- & 0 & 0.049 \\

\textbf{SSP-DMGTimeNet (ours)}
& \textbf{0.65}
& \textbf{1.457}
& \textbf{1.237}
& \textbf{1847}
& \textbf{0.00}
& \textbf{0.898}
& \textbf{62}
& 0.116 \\

\midrule
GT reference
& 93.55 & 9.683 & 118.349 & 62
& --- & --- & 62 & --- \\

\bottomrule
\end{tabular*}

\vspace{2pt}
\begin{minipage}{\textwidth}
\footnotesize
$^{*}$ Per-model criterion: predicted-leader detrended RMS $\geq 0.05$\,m/s; 1,847 test windows in total.
The GT-excitation criterion uses 62 windows with GT-leader detrended RMS $\geq 0.05$\,m/s.
$^{\dagger}$ Constant-speed leader; stability metrics are undefined.
GT reference row shown for context only and excluded from best-value comparison.
\end{minipage}

\end{table*}

\subsection{Zero-Shot Generalization on NGSIM}

To evaluate cross-dataset generalization, models trained exclusively on HighD are directly evaluated on NGSIM US-101 and I-80 without target-domain fine-tuning or re-estimation of normalization statistics. We sample 4,000 vehicle-platoon windows from each roadway using the same input and output configuration as in the HighD experiments. The results are reported in \Tabref{tab:ngsim-us101} and \Tabref{tab:ngsim-i80}.

SSP-DMGTimeNet achieves velocity MAEs of 1.316\,m/s on US-101 and 1.252\,m/s on I-80, remaining competitive among the learning-based models. By comparison, Int-LSTM, which achieves the lowest velocity error on HighD, degrades to 3.343\,m/s and 3.721\,m/s, respectively. This result suggests that SSP-DMGTimeNet is less dependent on the training-domain distribution.

The advantage is more pronounced in platoon-level stability. SSP-DMGTimeNet achieves unstable-window rates of 3.90\% and 4.10\% on US-101 and I-80, respectively, compared with 87.52--97.33\% for the other learning-based models. On windows with valid ground-truth leader excitation, the corresponding rates remain low at 4.12\% and 4.22\%, while SSP-DMGTimeNet covers all 3,665 US-101 and 3,485 I-80 excitation windows. These results indicate that the learned disturbance-suppression behavior transfers to traffic conditions not observed during training.

\begin{table*}[t]
\centering
\caption{Zero-shot evaluation on NGSIM US-101 using models trained on HighD.}
\label{tab:ngsim-us101}
\small
\resizebox{\textwidth}{!}{%
\begin{tabular}{lrrrrrrr}
\toprule
Model &
$v$-MAE$\downarrow$ &
Tail $v$-MAE &
Unstable (\%)$^{*}$ &
GT-subset unstable (\%) &
GT-subset max. amp. &
GT-subset valid/3665 &
RMS jerk \\
\midrule
Int-LSTM                     & 3.343 & 3.252 & 96.10 & 96.08 & 16.581 & 3396 & 1.037 \\
Transformer                  & 1.340 & 1.241 & 95.61 & 95.49 & 9.235  & 1243 & 0.505 \\
Full-graph Attention         & 1.321 & 1.190 & 90.69 & 90.87 & 10.022 & 1238 & 0.479 \\
LSTM                         & 2.333 & 1.962 & 88.72 & 88.55 & 6.359  & 3512 & 0.668 \\
IDM cascade$^{\dagger}$      & 0.811 & 0.798 & --- & --- & --- & 0 & 0.118 \\
CNN-Int-LSTM-IDM             & 1.286 & 1.183 & 92.74 & 92.77 & 390152.344 & 1507 & 1.288 \\
DMGTimeNet cascade$^{\dagger}$
                             & 1.299 & 1.439 & --- & --- & --- & 0 & 0.577 \\
OVM cascade$^{\dagger}$      & 0.856 & 0.846 & --- & --- & --- & 0 & 0.029 \\
FVDM cascade$^{\dagger}$     & 0.814 & 0.798 & --- & --- & --- & 0 & 0.040 \\
\textbf{SSP-DMGTimeNet (ours)}
                             & \textbf{1.316} & \textbf{1.213} & \textbf{3.90} &
                               \textbf{4.12} & \textbf{1.791} &
                               \textbf{3665} & \textbf{0.636} \\
\midrule
GT reference                 & --- & --- & --- & 98.55 & 290646.094 & 3665 & --- \\
\bottomrule
\end{tabular}%
}
\\[2pt]
{\footnotesize
$^{*}$ Per-model excitation floor: predicted-leader detrended RMS $\geq 0.05$\,m/s.
The GT subset additionally requires GT-leader excitation $\geq 0.05$\,m/s; 3,665 such windows are available on US-101.
$^{\dagger}$ Constant-speed leader; stability metrics are undefined.}
\end{table*}

\begin{table*}[t]
\centering
\caption{Zero-shot evaluation on NGSIM I-80 using models trained on HighD.}
\label{tab:ngsim-i80}
\small
\resizebox{\textwidth}{!}{%
\begin{tabular}{lrrrrrrr}
\toprule
Model &
$v$-MAE$\downarrow$ &
Tail $v$-MAE &
Unstable (\%)$^{*}$ &
GT-subset unstable (\%) &
GT-subset max. amp. &
GT-subset valid/3485 &
RMS jerk \\
\midrule
Int-LSTM                     & 3.721 & 3.823 & 96.27 & 96.21 & 20.399 & 3142 & 1.028 \\
Transformer                  & 1.277 & 1.247 & 97.33 & 97.29 & 8.499  & 921  & 0.504 \\
Full-graph Attention         & 1.256 & 1.182 & 88.71 & 88.54 & 9.407  & 960  & 0.474 \\
LSTM                         & 2.148 & 1.735 & 87.52 & 86.83 & 5.004  & 3310 & 0.644 \\
IDM cascade$^{\dagger}$      & 0.781 & 0.811 & --- & --- & --- & 0 & 0.178 \\
CNN-Int-LSTM-IDM             & 1.108 & 1.049 & 93.95 & 93.99 & 262334.781 & 1298 & 1.363 \\
DMGTimeNet cascade$^{\dagger}$
                             & 1.220 & 1.411 & --- & --- & --- & 0 & 0.557 \\
OVM cascade$^{\dagger}$      & 0.863 & 0.912 & --- & --- & --- & 0 & 0.031 \\
FVDM cascade$^{\dagger}$     & 0.847 & 0.892 & --- & --- & --- & 0 & 0.045 \\
\textbf{SSP-DMGTimeNet (ours)}
                             & \textbf{1.252} & \textbf{1.184} & \textbf{4.10} &
                               \textbf{4.22} & \textbf{1.536} &
                               \textbf{3485} & \textbf{0.581} \\
\midrule
GT reference                 & --- & --- & --- & 98.31 & 443271.375 & 3485 & --- \\
\bottomrule
\end{tabular}%
}
\\[2pt]
{\footnotesize
$^{*}$ The evaluation criterion follows \Tabref{tab:ngsim-us101}; 3,485 GT-excitation windows are available on I-80.
$^{\dagger}$ Constant-speed leader; stability metrics are undefined.}
\end{table*}

\subsection{Sensitivity Analysis on Platoon Length}

To assess the robustness of SSP-DMGTimeNet to varying platoon sizes, we independently construct and train models for platoons with $N=3$, 5, 6, and 7 vehicles.

\begin{figure*}
\centering
\includegraphics[width=\linewidth]{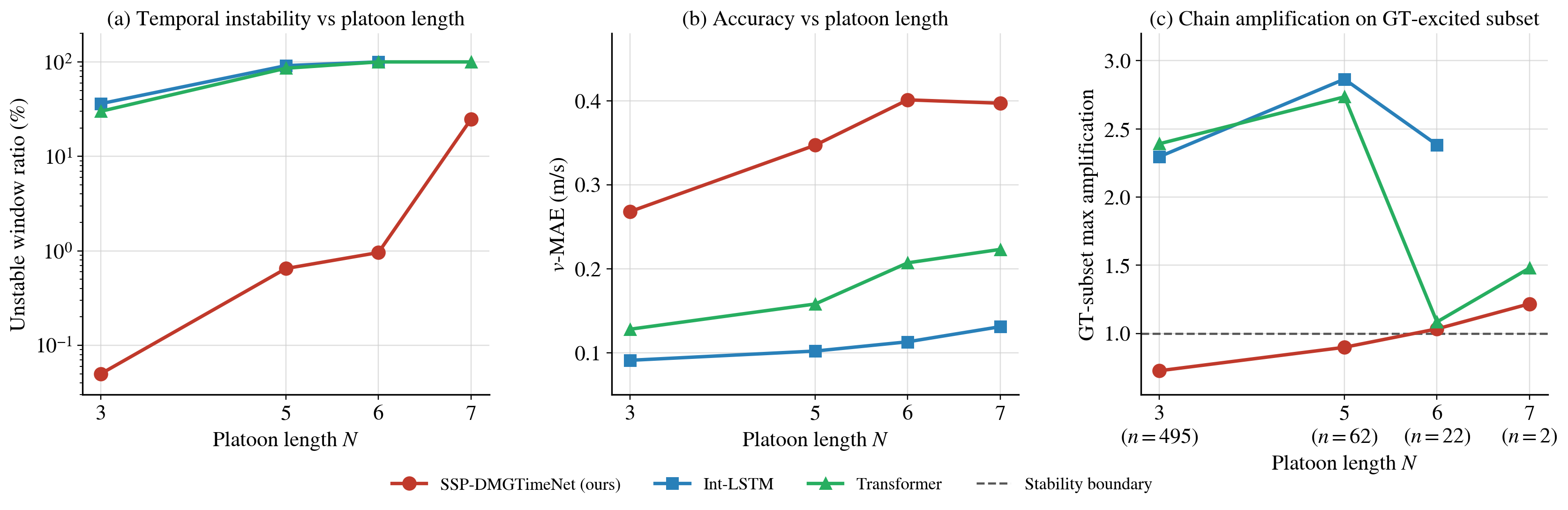}
\caption{Performance under different platoon lengths ($N=3,5,6,7$): (a) unstable-window rate, (b) velocity MAE, and (c) maximum amplification on the GT-excitation subset.}
\label{fig:queue_length}
\end{figure*}

As shown in \Figref{fig:queue_length}(a), SSP-DMGTimeNet achieves unstable-window rates of 0\%, 0.65\%, and 0.96\% for $N=3$, 5, and 6, respectively, remaining below 1\% and substantially lower than Int-LSTM and Transformer. When the platoon length increases to $N=7$, the unstable-window rate rises to 24.73\%, indicating that disturbance suppression becomes more difficult over longer propagation chains, although it remains markedly lower than the 100\% observed for Transformer. \Figref{fig:queue_length}(b) shows that the velocity MAE of SSP-DMGTimeNet remains within 0.268--0.401\,m/s across all platoon lengths, without systematic growth as $N$ increases.

\Figref{fig:queue_length}(c) further reports the maximum amplification on the GT-excitation subset. SSP-DMGTimeNet yields values of 0.726 and 0.898 for $N=3$ and $N=5$, respectively, both below the string-stability boundary of one. The values increase to 1.033 and 1.217 for $N=6$ and $N=7$, indicating the emergence of localized disturbance amplification in longer platoons. Overall, SSP-DMGTimeNet maintains robust disturbance-suppression performance up to moderate platoon lengths, with degradation becoming more evident as the propagation chain increases.

\subsection{Ablation Study}

We conduct ablation experiments by removing individual architectural components or loss terms from the full model. All variants are trained under the same settings and evaluated on the HighD test set. Stability metrics on the GT-excitation subset are computed using the same 62 reference windows.

\begin{table*}[t]
\centering
\caption{Ablation results on HighD with stability evaluated on 62 GT-excitation windows.}
\label{tab:ablation}
\small
\begin{tabular*}{\textwidth}
{@{\extracolsep{\fill}}lccccccc@{}}
\toprule
Variant &
$v$-MAE &
\makecell{Unstable\\(\%)$^{*}$} &
\makecell{Max.\\amp.$^{*}$} &
\makecell{Exceedance\\area$^{*}$} &
\makecell{Max. freq.\\gain} &
\makecell{GT unstable\\(\%)} &
\makecell{GT max.\\amp.} \\
\midrule
SSP (full)           & 0.347 & 0.65 & 1.457 & 1.237  & 2.165  & 0.00 & 0.898 \\
Full graph           & 0.372 & 0.49 & 1.256 & 0.705  & 2.593  & 0.00 & 0.926 \\
Fixed $\tau$         & 0.375 & 0.97 & 1.244 & 1.375  & 2.037  & 1.61 & 1.101 \\
No delay bias        & 0.382 & 0.38 & 1.328 & 0.680  & 2.111  & 0.00 & 0.896 \\
No CFE               & 0.385 & 0.22 & 1.184 & 0.270  & 1.832  & 0.00 & 0.936 \\
No HGF               & 0.371 & 0.60 & 1.306 & 1.170  & 3.746  & 1.61 & 1.036 \\
No $L_{\mathrm{adj}}$ & 0.336 & 0.43 & 1.360 & 0.796  & 2.170  & 0.00 & 0.973 \\
No $L_{\mathrm{fft}}$ & 0.303 & 3.74 & 1.733 & 10.332 & 10.208 & 8.06 & 1.103 \\
No $L_{\mathrm{sub}}$ & 0.366 & 0.11 & 1.073 & 0.081  & 2.237  & 0.00 & 0.864 \\
\bottomrule
\end{tabular*}

\vspace{2pt}
{\footnotesize
$^{*}$ Computed only for windows with predicted-leader detrended RMS $\geq 0.05$\,m/s.}
\end{table*}

\textbf{Disturbance-propagation mechanism.}
Replacing the directional spatial causal mask with full-graph attention increases the velocity MAE from 0.347 to 0.372 and the maximum frequency-domain gain from 2.165 to 2.593, indicating that restricting information flow to the upstream-to-downstream direction helps preserve physically consistent inter-vehicle interactions. Fixing the propagation delay increases the GT-subset maximum amplification from 0.898 to 1.101 and produces a 1.61\% unstable-window rate. These results support learning the propagation delay directly from data rather than imposing a fixed value.

\textbf{Feature representations.}
Removing HGF increases the GT-subset maximum amplification to 1.036 and the maximum frequency-domain gain to 3.746, suggesting that adaptive temporal-scale fusion contributes to stable trajectory prediction. Removing CFE increases the velocity MAE to 0.385, the largest accuracy degradation among the feature-ablation variants, confirming the value of cross-vehicle cointegration residuals for representing inter-vehicle dynamics.

\textbf{String-stability regularization.}
Removing $L_{\mathrm{fft}}$ reduces the velocity MAE to 0.303 but increases the GT-subset unstable-window rate to 8.06\%, the maximum amplification to 1.103, and the maximum frequency-domain gain to 10.208. This result reveals a clear accuracy--stability trade-off and highlights the role of frequency-domain regularization in suppressing selective disturbance amplification.

Overall, the ablation results show that the major components play complementary roles. SP-DACA captures the direction and delay of disturbance propagation, HGF and CFE improve temporal and cross-vehicle representations, and the string-stability losses directly suppress disturbance amplification. The full model therefore provides a balanced compromise between trajectory accuracy and time- and frequency-domain stability.

\subsection{Interpretability Analysis}

We analyze the learned physical structure of SSP-DMGTimeNet from three complementary perspectives: learned propagation delays, consistency between predicted and observed delays, and disturbance propagation in the time and frequency domains.

\subsubsection{Learned Propagation Delays}

\begin{figure}[!t]
\centering
\includegraphics[width=\linewidth]{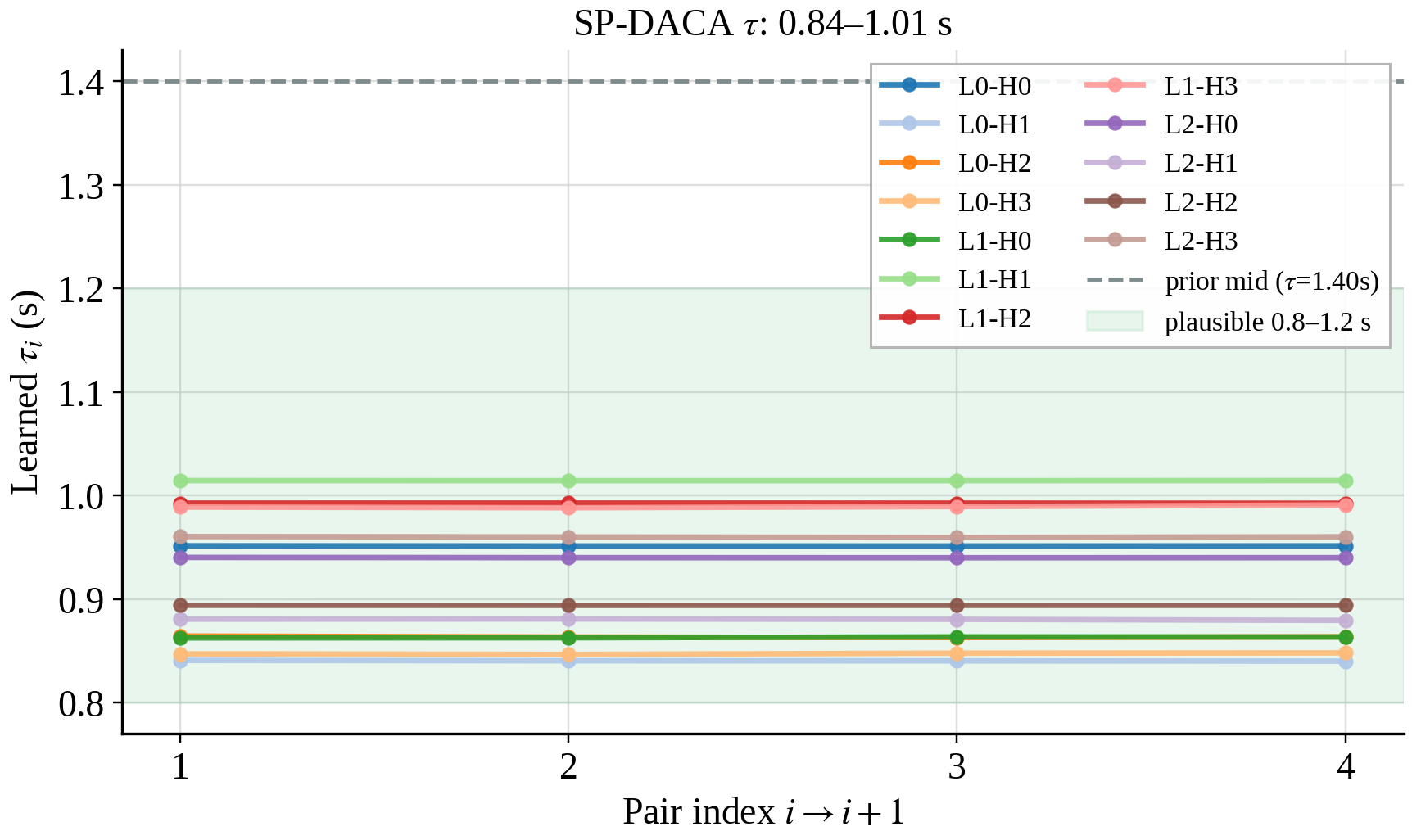}
\caption{Learned propagation delays in SP-DACA. The red dashed line indicates the initialization midpoint of 1.40\,s, and the shaded region denotes the physically plausible range of 0.8--1.2\,s.}
\label{fig:interp_a}
\end{figure}

\Figref{fig:interp_a} shows the propagation delays learned by SP-DACA across three attention layers with four heads per layer. The red dashed line denotes the initialization midpoint, $\tau=1.40$\,s, while the shaded region indicates the physically plausible range of 0.8--1.2\,s associated with driver response time and stable time headway.

The learned delays range from 0.84 to 1.01\,s, substantially below the initialization midpoint and entirely within the plausible interval. This convergence indicates that the delay parameters are effectively updated during training rather than remaining near their initialization. The attention heads also exhibit distinct delay values within this range, with a maximum difference of approximately 0.17\,s.

\subsubsection{Consistency of Predicted Propagation Delays}

\begin{figure}[!t]
\centering
\includegraphics[width=\linewidth]{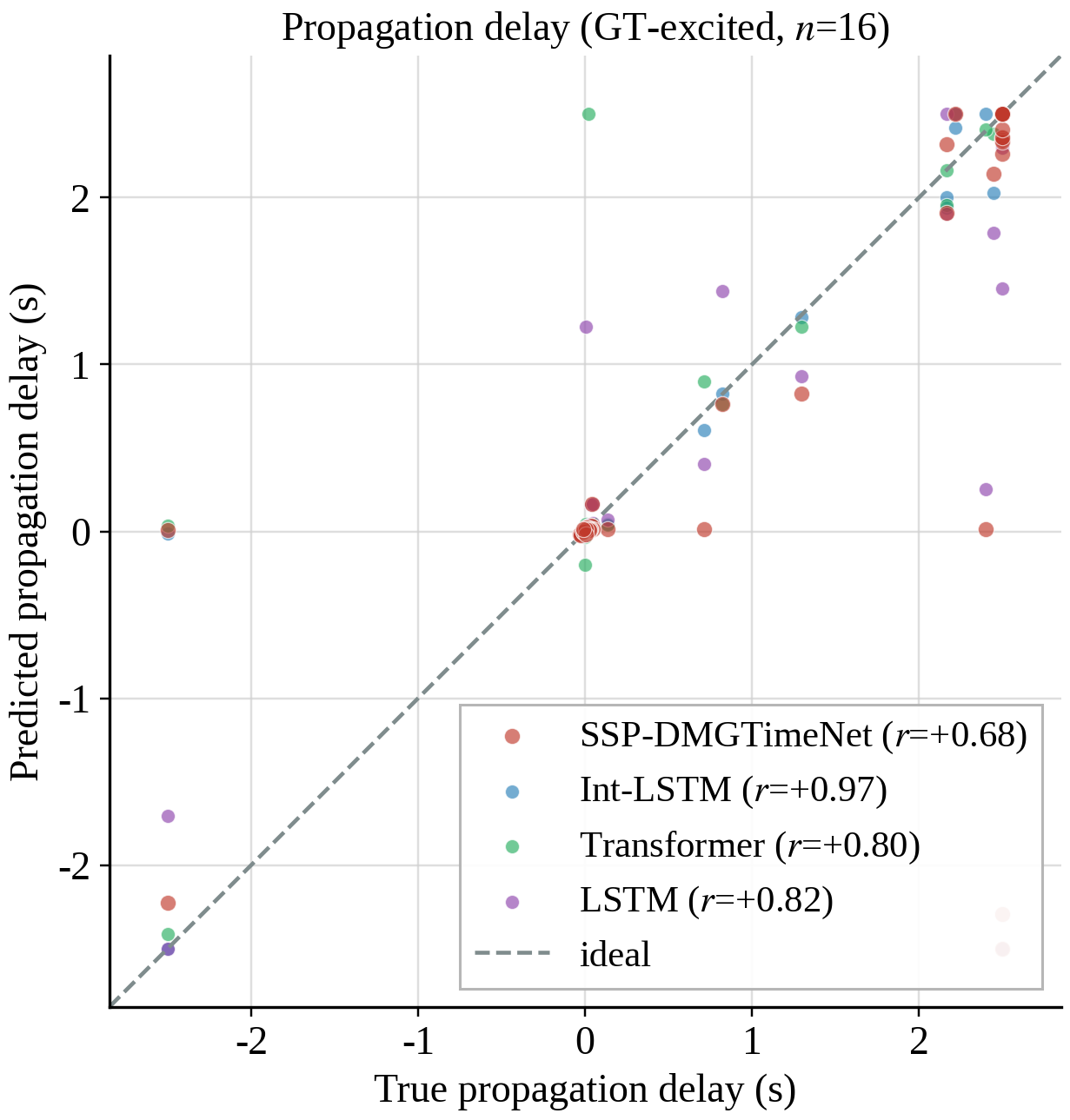}
\caption{Consistency between ground-truth and predicted propagation delays. The dashed line $y=x$ indicates ideal recovery.}
\label{fig:interp_b}
\end{figure}

To assess whether the predicted trajectories preserve the temporal structure of disturbance propagation, we estimate adjacent-vehicle propagation delays from both ground-truth and predicted trajectories using cross-correlation peaks, as shown in \Figref{fig:interp_b}. The analysis uses the concatenated 8\,s historical and predicted trajectories with a maximum admissible delay of 2.5\,s. To avoid unreliable estimates in near-stationary windows, we retain 66 windows with GT-leader disturbance amplitudes above 0.05\,m/s, yielding 264 adjacent vehicle pairs.

SSP-DMGTimeNet achieves a Pearson correlation of $r=0.58$ between the ground-truth and predicted propagation delays, indicating a moderate recovery of the temporal propagation structure. For ground-truth delays within approximately 0.8--1.2\,s, the predictions are concentrated near the ideal diagonal. Larger dispersion occurs for near-zero delays, where cross-correlation-based estimation is intrinsically less reliable.

\subsubsection{Physical Structure of Disturbance Propagation and String Stability}

\begin{figure}[!t]
  \centering
  \includegraphics[width=\linewidth]{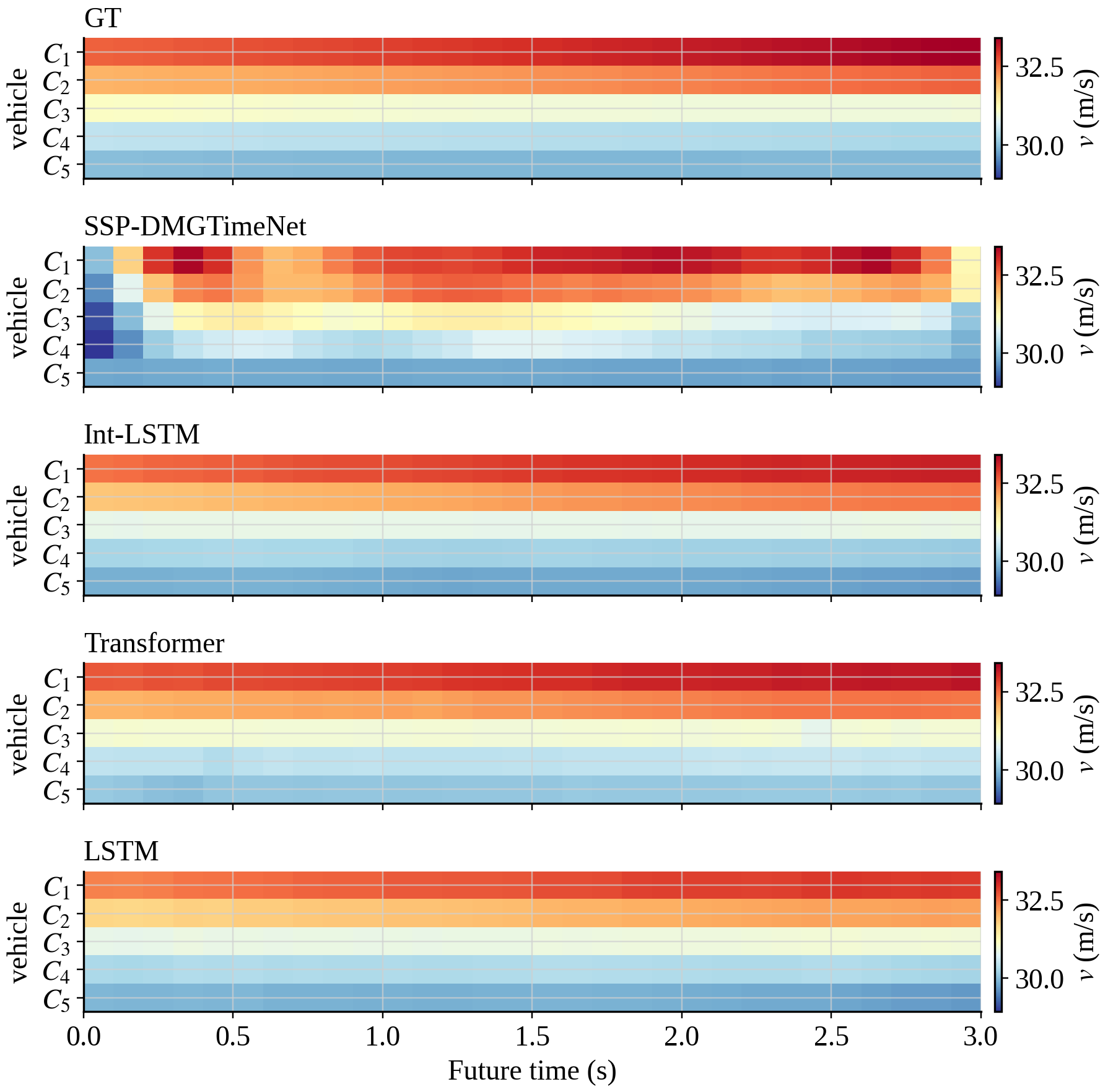}
  \caption{Velocity-disturbance heatmaps of the ground truth and each model.
  The diagonal pattern reflects disturbance propagation from the leader ($i=0$)
  to the tail vehicle ($i=4$).}
  \label{fig:vel_heatmap}
  \end{figure}

  We further examine whether SSP-DMGTimeNet preserves the physical propagation
  structure of disturbances while suppressing amplification. As shown in
  \Figref{fig:vel_heatmap}, the ground-truth trajectories exhibit a diagonal
  velocity-disturbance pattern propagating from the leader ($i=0$) toward the
  tail vehicle ($i=4$). SSP-DMGTimeNet reproduces a similar propagation pattern,
  preserving the arrival order and attenuation trend without evident amplitude
  amplification. In comparison, Int-LSTM and LSTM produce less distinct
  disturbance boundaries, while Transformer exhibits noticeable deviations for
  intermediate vehicles.

\begin{figure}[!t]
  \centering
  \includegraphics[width=\linewidth]{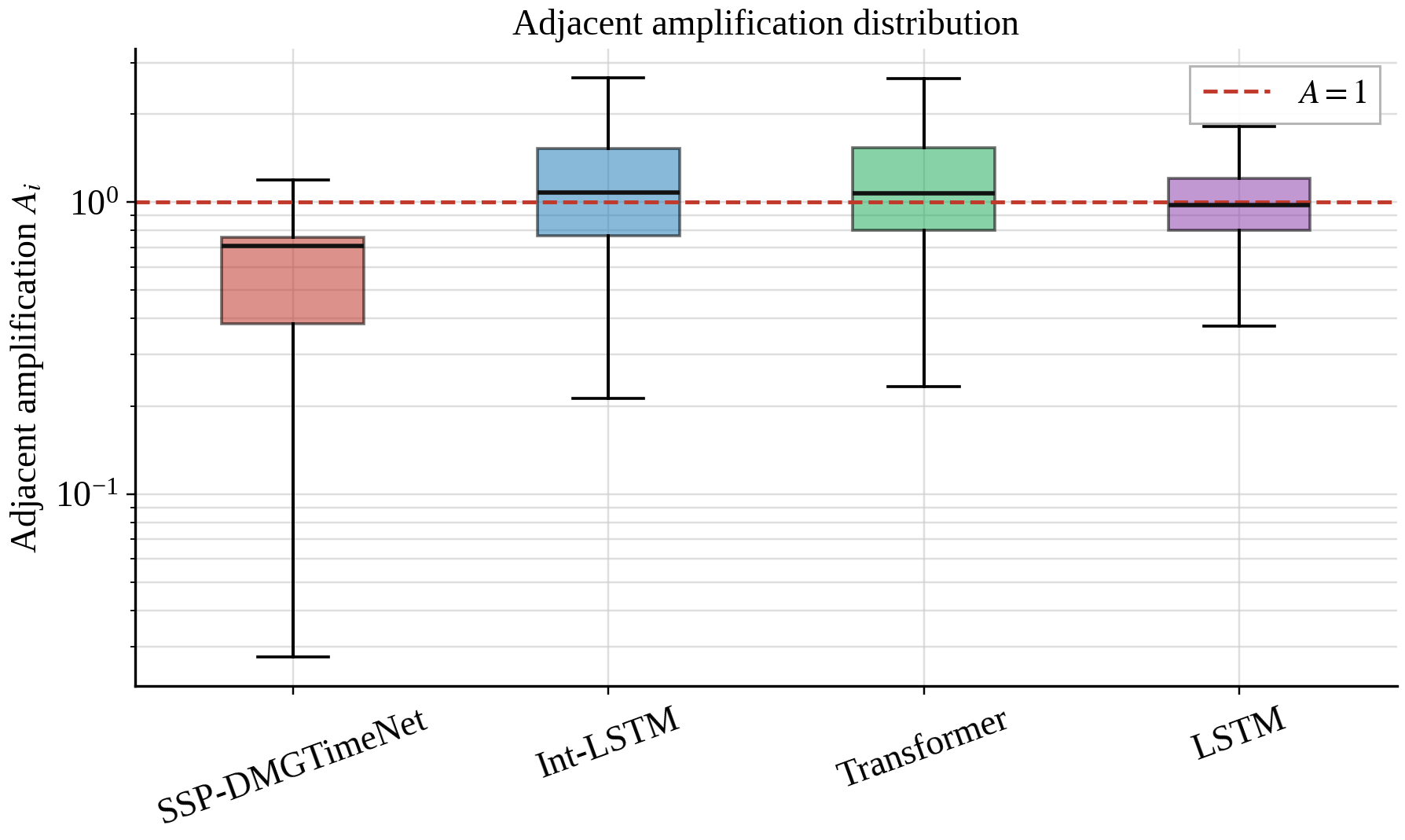}
  \caption{Distributions of adjacent-vehicle amplification factors $A_i$.
  Values below one (dashed line) indicate string-stable attenuation.}
  \label{fig:amp_box}
  \end{figure}

  Beyond the qualitative propagation pattern, \Figref{fig:amp_box} quantifies
  string stability through the adjacent-vehicle amplification factors $A_i$.
  SSP-DMGTimeNet has a median value of approximately 0.6 with a compact
  distribution and only a few outliers. In contrast, the distributions of
  Int-LSTM, Transformer, and LSTM are centered close to or above the stability
  boundary of one, with their upper quartiles extending beyond it.

\begin{figure}[!t]
  \centering
  \includegraphics[width=\linewidth]{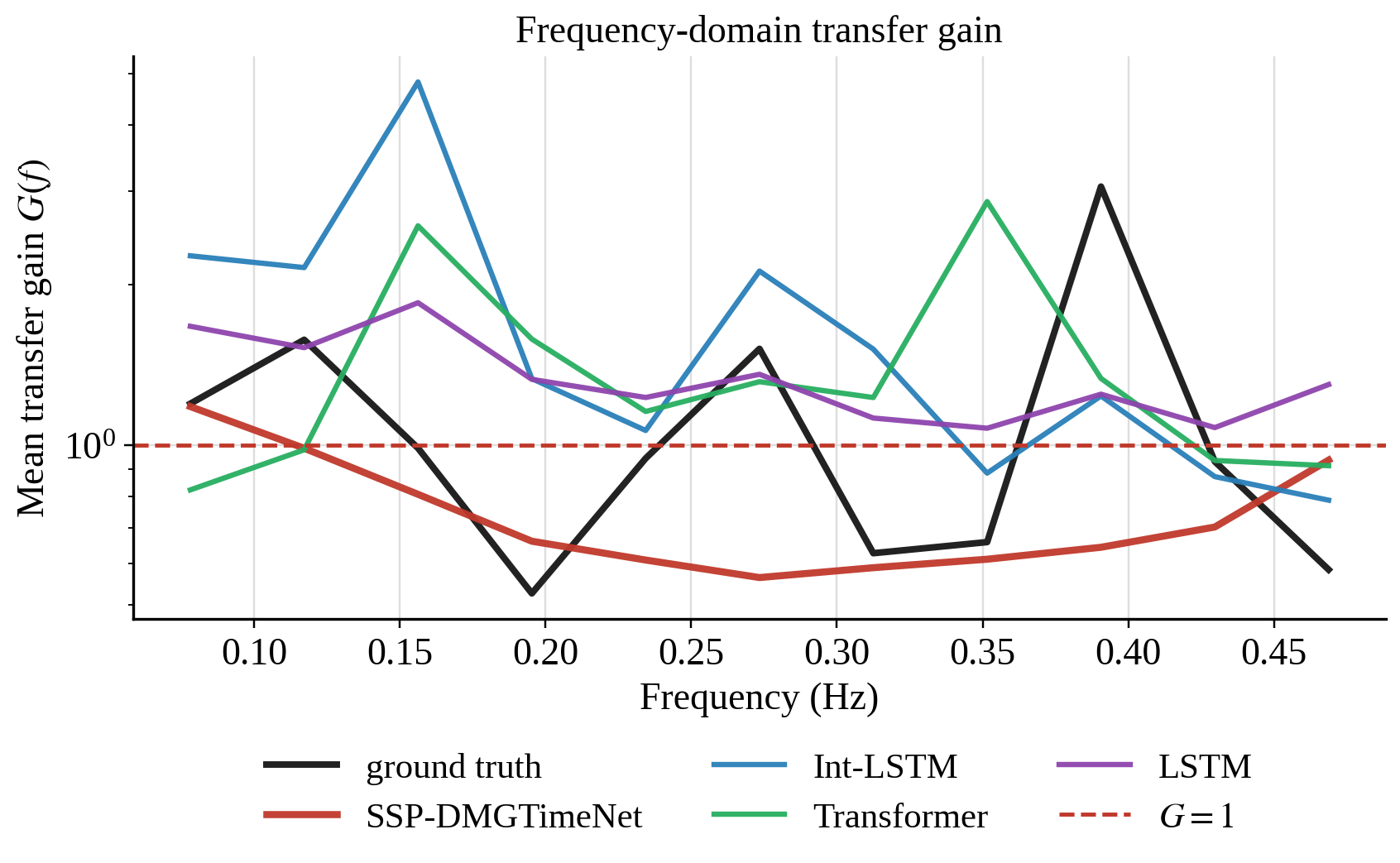}
  \caption{Frequency-domain transfer gains $G(f)$ over 0.05--0.5\,Hz. Gains
  above one indicate frequency-selective disturbance amplification.}
  \label{fig:fft_gain}
  \end{figure}

  The frequency-domain transfer gains in \Figref{fig:fft_gain} provide a
  complementary view.Across 0.05--0.5\,Hz, SSP-DMGTimeNet maintains transfer
  gains of approximately 0.5--0.9 without frequency-selective amplification.
  Int-LSTM exhibits a gain above two near 0.1\,Hz, Transformer shows a secondary
  increase around 0.35\,Hz, and LSTM fluctuates around the stability boundary.
  Together, these results indicate that SSP-DMGTimeNet preserves the temporal
  structure of disturbance propagation while suppressing amplification in both
  the time and frequency domains.

\section{Conclusion}

This paper presents SSP-DMGTimeNet, a physics-constrained learning framework for spatiotemporal trajectory prediction of vehicle platoons. The framework incorporates disturbance propagation through propagation-delay-aware causal attention, cross-vehicle representations, and time- and frequency-domain string-stability constraints. On HighD, SSP-DMGTimeNet achieves an unstable-window rate of 0.65\% and a maximum head-to-tail amplification of 0.898 on the GT-excitation subset, while maintaining competitive prediction performance. The learned propagation delays converge to a physically plausible range of 0.84--1.01\,s, supporting the model's ability to capture inter-vehicle disturbance transmission. Zero-shot evaluation on NGSIM US-101 and I-80 yields unstable-window rates of 3.90\% and 4.10\%, respectively, demonstrating cross-dataset stability. Experiments with different platoon lengths further show that disturbance suppression remains effective for moderate platoon sizes but becomes more challenging as the propagation chain increases. Overall, these results show that string stability can serve as a training-time physical constraint that complements conventional trajectory objectives with platoon-level dynamic consistency. Future work will extend SSP-DMGTimeNet to heterogeneous platoons and evaluate its long-horizon performance in closed-loop CACC and MPC systems.

\bibliography{related_work_references}
\bibliographystyle{unsrtnat}

\section*{Acknowledgements}
This work was supported by the Korea Institute of Science and Technology Information (KISTI) under the R\&D program ``Development of the Next-Generation Integrated Wired/Wireless Communication Gateway (X-Gateway).'' 

% ===============================================================
% Author biographies
% ===============================================================

% ===============================================================
% Author biographies
% ===============================================================

%\authorbio{photo/yuhang.jpg}{Yuhang Wang}{%received the Ph.D. degree from Tsinghua University in 2023. She was a Visiting Scholar with the Department of Cognitive Robotics, Delft University of Technology, from 2021 to 2022, and a Postdoctoral Associate with the MIT SMART Center and the University of Wisconsin--Madison. She is currently an Assistant Professor with the Cho Chun Shik Graduate School of Mobility, KAIST. Her research interests include safe and trustworthy autonomy, generative AI, and human-centered AI for autonomous systems.}

\authorbio{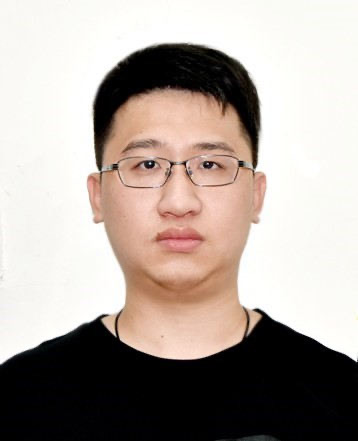}{Yuhang Wang}{Yuhang Wang received the B.S. degree from Shenyang University of Technology (SUT), Shenyang, China, in 2019, and the M.S. degree from the University of Chinese Academy of Sciences (UCAS), Beijing, China, in 2025. He is currently with the China National Petroleum Corporation (CNPC), where he is responsible for information security. He is also a Collaborative Researcher with the group led by Prof. Heye Huang at the Korea Advanced Institute of Science and Technology (KAIST), Daejeon, South Korea. His research interests include trajectory generation, multi-sensor fusion, safety-critical scenarios for automated vehicles, and information security.}

\authorbio{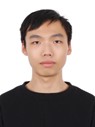}{Kailang Ma}{received the B.S. degree in information security and the M.S. degree in cyberspace security from Beihang University, Beijing, China, in 2021 and 2024, respectively. He was an Algorithm Engineer at WeRide, an autonomous driving company, from 2024 to 2026. He is currently pursuing the Ph.D. degree with the Cho Chun Shik Graduate School of Mobility, Korea Advanced Institute of Science and Technology (KAIST), under the supervision of Prof. Heye Huang. His research interests include safe and trustworthy autonomy, generative AI for autonomous driving, world models, and risk-sensitive decision making.}

\authorbio{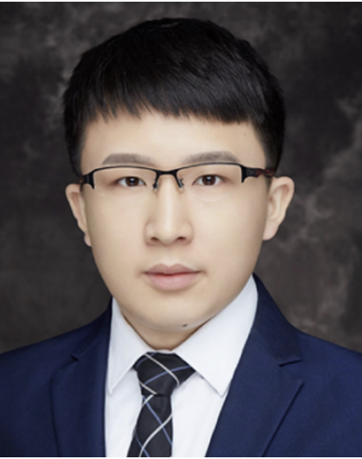}{Zirui Li}{received the B.S. and Ph.D. degrees in mechanical engineering from Beijing Institute of Technology (BIT), Beijing, China, in 2019 and 2025, respectively. From June 2021 to July 2022, he was a Visiting Researcher with Delft University of Technology (TU Delft), Delft, The Netherlands. From August 2022 to June 2024, he was a Visiting Researcher with the Chair of Traffic Process Automation, Faculty of Transportation and Traffic Sciences ``Friedrich List,'' TU Dresden, Dresden, Germany. He is currently a Postdoctoral Research Fellow with Nanyang Technological University, Singapore. His research interests include interactive behavior modeling, risk assessment, and motion planning for automated vehicles.
}

\authorbio{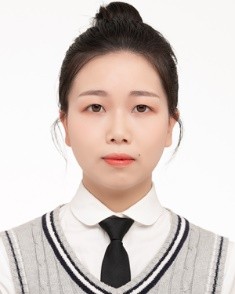}{Mingfeng Fan}{received the B.S. and Ph.D. degrees in traffic and transportation engineering from Central South University, Changsha, China, in 2019 and 2024, respectively. She is currently a Research Fellow with the Department of Mechanical Engineering, National University of Singapore. Her research interests include deep reinforcement learning, combinatorial optimization, and robotic control and scheduling.
}

\authorbio{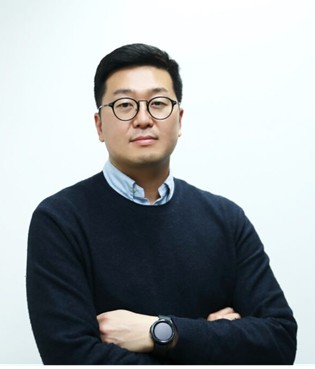}{Kitae Jang}{received the Ph.D. degree in civil and environmental engineering from the University of California, Berkeley, in 2011. He is currently a Professor with the Cho Chun Shik Graduate School of Mobility, KAIST. He also serves as the Dean of KAIST Academy and the Director of the Center for Excellence in Learning and Teaching and the Research Center for Eco-friendly and Smart Vehicles. His current research interests include transportation energy and environmental policy, traffic operations, intelligent transportation systems, and transportation systems for healthy and safe communities.
}

\authorbio{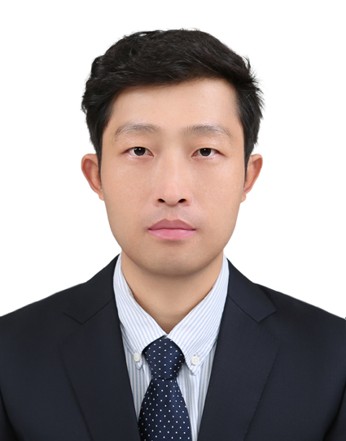}{Changju Lee}{received his Ph.D. in Civil Engineering, with a focus on transportation planning, from the University of Virginia in 2015. He is currently an Associate Professor at the Cho Chun Shik Graduate School of Mobility at KAIST. Prior to joining KAIST, he worked at the United Nations Economic and Social Commission for Asia and the Pacific (UN ESCAP), the Virginia Transportation Research Council, and LG CNS Co., Ltd. His research interests include sustainable transportation policy, smart and low-carbon mobility planning, transportation and climate change, applications of big data and machine learning, and urban and transportation economics.
}

\authorbio{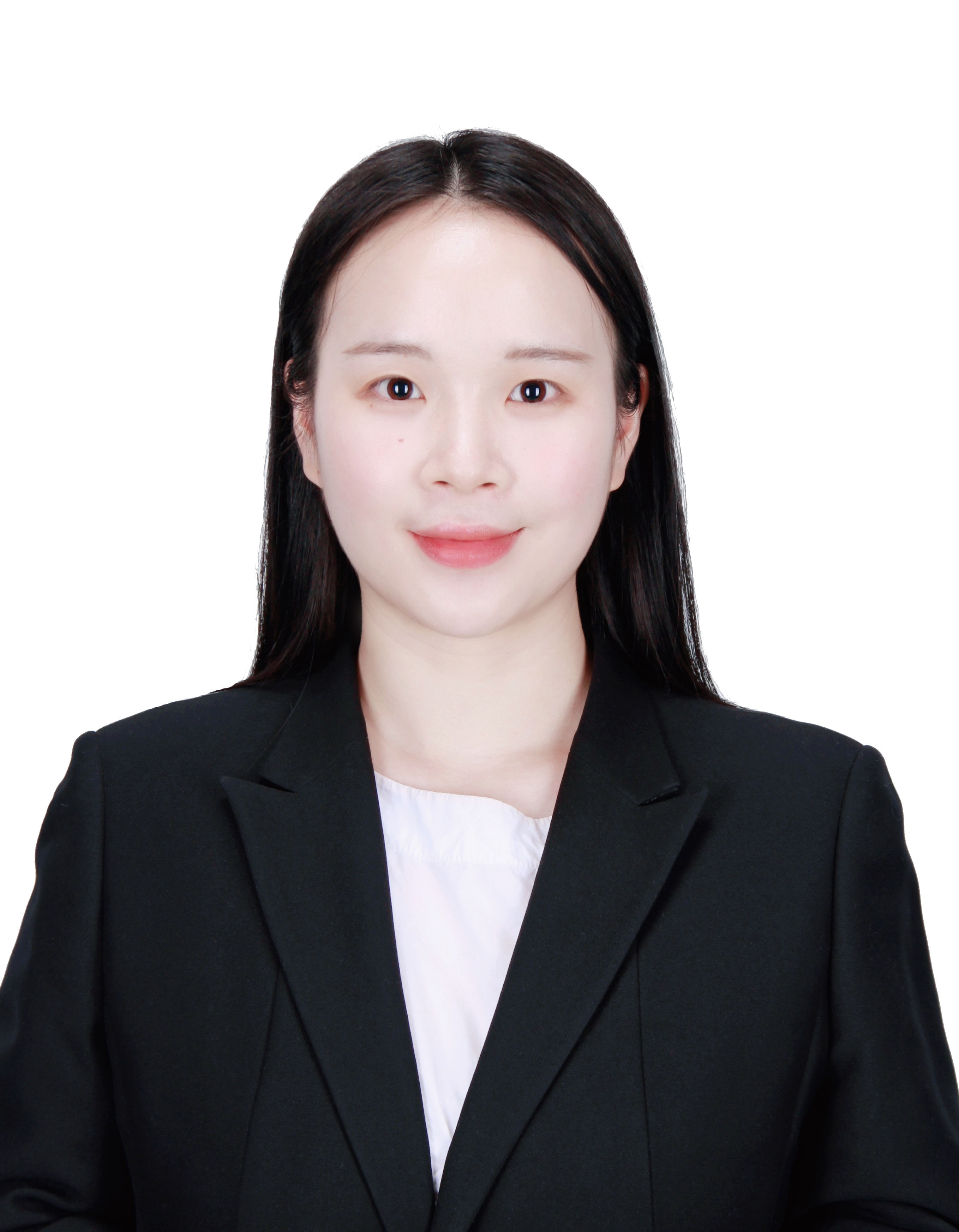}{Heye Huang}{received the Ph.D. degree from Tsinghua University in 2023. He was a Visiting Scholar with the Department of Cognitive Robotics, Delft University of Technology, from 2021 to 2022, and a Postdoctoral Associate with the MIT SMART Center and the University of Wisconsin--Madison. He is currently an Assistant Professor with the Cho Chun Shik Graduate School of Mobility, KAIST. His research interests include safe and trustworthy autonomy, generative AI, and human-centered AI for autonomous systems.
}

\end{document}